\DocumentMetadata{}
\documentclass[acmlarge,regular,solid]{acmart}

\usepackage{lscape}
\usepackage{array}
\usepackage[utf8]{inputenc}
\usepackage{graphicx}
\usepackage[table]{xcolor} % in preamble
\usepackage{longtable} %% Addtion to the old one
\usepackage{tikz} %% Addtion to the old one
\usepackage{lipsum} %% Addtion to the old one

\usepackage{forest,xcolor}
\usepackage{fontawesome5} %
\usepackage{enumitem}

\usepackage{graphicx}

\usetikzlibrary{shapes, arrows}

\usepackage{tikz}
\usetikzlibrary{arrows.meta, positioning, shapes.multipart}

\usepackage[autostyle]{csquotes}
\AtBeginDocument{%
  \providecommand\BibTeX{{%
    \normalfont B\kern-0.5em{\scshape i\kern-0.25em b}\kern-0.8em\TeX}}}

\setcopyright{acmcopyright}
\copyrightyear{2025}
\acmYear{2025}
\acmDOI{XXXXXXX.XXXXXXX}

\acmJournal{POMACS}
\acmVolume{XX}
\acmNumber{X}
\acmArticle{XXX}
\acmMonth{5}
\begin{document}

% \title {LLM Agents for Data Science: A Survey of Taxonomy, Capabilities, and Trust
% }

\title{LLM-Based Data Science Agents: A Survey of Capabilities, Challenges, and Future Directions}

\author{Mizanur Rahman}
% \authornotemark[1]
\affiliation{%
  \institution{York University}
  \city{Toronto}
  \country{Canada}}
\email{mizanurr@yorku.ca}

\author{Amran Bhuiyan}
\authornote{Equal contribution}
% \authornotemark[1]
\affiliation{%
  \institution{York University}
  \country{Canada}}
\email{amran.apece@gmail.com}

% \author{Mohammed Saidul Islam}
% \authornotemark[1]
% \affiliation{%
%   \institution{York University}
%   \country{Canada}}
% \email{saidulis@yorku.ca}

\author{Mohammed Saidul Islam}
\authornotemark[1]
\affiliation{%
  \institution{York University}
  \country{Canada}}
\affiliation{%
  \institution{Vector Institute for AI}
  \country{Canada}}
\email{saidulis@yorku.ca}

% \author{Md Tahmid Rahman Laskar}
% \authornotemark[1]
% \affiliation{%
%   \institution{York University}
%   \country{Canada}}
% \email{tahmid20@yorku.ca}

\author{Md Tahmid Rahman Laskar}
\authornotemark[1]
\affiliation{%
  \institution{York University}
  \country{Canada}}
\affiliation{%
  \institution{Dialpad Inc.}
  \country{Canada}}
\email{tahmid20@yorku.ca}

\author{Ridwan Mahbub}
\authornotemark[1]
\affiliation{%
  \institution{York University}
  \country{Canada}}
\email{rmahbub@yorku.ca}
\author{Ahmed Masry}
\authornotemark[1]
\affiliation{%
  \institution{York University}
  \country{Canada}}
\email{masry20@yorku.ca}

% \author{Shafiq Joty}
% \affiliation{%
%   \institution{Salesforce AI Research}
%   \country{USA}}
% \email{shafiqrayhan@gmail.com }

\author{Shafiq Joty}
\affiliation{%
  \institution{Nanyang Technological University}
  \country{Singapore}}
\affiliation{%
  \institution{Salesforce AI Research}
  \country{USA}}
\email{shafiqrayhan@gmail.com }

\author{Enamul Hoque}
\affiliation{%
  \institution{York University}
  \country{Canada}}
\email{enamulh@yorku.ca}

% \author{Jimmy Xiangji Huang}
% \affiliation{%
%   \institution{York University}
%   \city{Toronto}
%   \country{Canada}}
% \email{jhuang@yorku.ca}

\renewcommand{\shortauthors}{Rahman, Enamul, et al.}

\begin{abstract}
Recent advances in large language models (LLMs) have enabled a new class of AI agents that automate multiple stages of the data science workflow by integrating planning, tool use, and multimodal reasoning across text, code, tables, and visuals. This survey presents the first comprehensive, lifecycle-aligned taxonomy of data science agents, systematically analyzing and mapping forty-five systems onto the six stages of the end-to-end data science process: business understanding and data acquisition, exploratory analysis and visualization, feature engineering, model building and selection, interpretation and explanation, and deployment and monitoring. In addition to lifecycle coverage, we annotate each agent along five cross-cutting design dimensions: reasoning and planning style, modality integration, tool orchestration depth, learning and alignment methods, and trust, safety, and governance mechanisms. Beyond classification, we provide a critical synthesis of agent capabilities, highlight strengths and limitations at each stage, and review emerging benchmarks and evaluation practices. Our analysis identifies three key trends: most systems emphasize exploratory analysis, visualization, and modeling while neglecting business understanding, deployment, and monitoring; multimodal reasoning and tool orchestration remain unresolved challenges; and over 90\% lack explicit trust and safety mechanisms. We conclude by outlining open challenges in alignment stability, explainability, governance, and robust evaluation frameworks, and propose future research directions to guide the development of robust, trustworthy, low-latency, transparent, and broadly accessible data science agents.
\end{abstract}




\newcommand{\joty}[1]{\textcolor{cyan}{(Joty: #1)}}
\newcommand{\enamul}[1]{\textcolor{blue}{(Enamul: #1)}}
\newcommand{\saidul}[1]{\textcolor{green}{(Saidul: #1)}}

\maketitle

\section{Introduction}
\label{sec:introduction}
% \enamul{major comments: 1) the paper needs to provide more gentle introductions to term and coherent connections between different parts of the survey. Right now it seems to cover too many topics with seemingly weak connections, if necessary even reduce the scope of the survey but make it more coherent. 2) Too text heavy, add more visuals, illustrative examples from different papers or real world examples, rather than dense texts and verbose tables. }
% \enamul{the introduction is little too short. It needs to gently introduce the keywords mentioned in the surveys  such as trust and safety, capabilities etc.}
The growing demand for data-driven decision-making across industries has made data science an essential capability \cite{cao2017data,sarker2021data}. However, realizing its full potential remains challenging due to steep skill barriers and the inherent complexity of end-to-end analytical workflows. Effective data science requires the seamless integration of expertise in business understanding, statistical analysis, data engineering, modeling, deployment, and visualization \cite{sarker2021data}, a combination of skills rarely present within a single individual or even across many teams. This persistent expertise gap continues to limit the scalable adoption of data-driven practices, particularly among non-technical users who struggle to navigate the full analytical lifecycle from problem formulation to production deployment. 

Recent advances in Large Language Models (LLMs) are reshaping the landscape of data science by expanding their capabilities far beyond natural language understanding \cite{yang2024harnessing,yu2024case}. LLMs can ingest raw data, generate visualizations, perform statistical analyses, build predictive models, and produce deployment-ready code, all through natural language interactions \cite{cheng2023gpt,hassan2023chatgpt}. These capabilities lower the entry barrier to advanced analytics and enable users without programming or machine learning expertise to engage with complex data science workflows. Leveraging these capabilities, researchers are increasingly developing autonomous agents, widely regarded as a pathway toward Artificial General Intelligence (AGI), that can reason, plan, and execute multi-step tasks. These agents are being applied across diverse domains, including software development \cite{hong2023metagpt}, data science \cite{zhang2024benchmarking}, robotics \cite{kannan2024smart}, societal simulation \cite{guo2024large}, policy modeling \cite{qian2024scaling}, gaming \cite{plaat2025agentic,wang2024survey}, scientific discovery \cite{novikov2025alphaevolve}, and advanced mathematical problem solving at the International Mathematical Olympiad level \cite{castelvecchi2025_imo}. Unlike traditional static systems, these agents decompose high-level objectives into subtasks, dynamically select and coordinate tools, and adapt their behavior based on intermediate feedback, thereby managing complex evolving, and long-horizon analytical workflows.

A growing number of these agents are now being designed specifically for data science tasks, known as data science (DS) agents. Centered on LLMs and extended with external tools, they aim to support the full data science lifecycle, from business understanding and data acquisition to analysis, modeling, evaluation, and deployment \cite{plaat2025agentic,wang2024survey,guo2024large,cheng2024exploring,qiao2023taskweaver}. Unlike general-purpose agents, DS agents are tailored to handle structured, high-dimensional data, apply robust statistical and machine learning methods, and coordinate tools to complete analytical workflows by retrieving data from heterogeneous databases or repositories, writing and executing reproducible code in notebooks, performing analyses, creating visualizations, and compiling reports. Their central goal is to enable users to specify high-level objectives in natural language while autonomously orchestrates the subsequent steps of collecting data, running analyses, training models, generating visualizations, and presenting results \cite{hassan2023chatgpt,yu2024case}. Early systems such as HuggingGPT \cite{shen2023hugginggpt}, GPT-4 as a Data Analyst \cite{cheng2023gpt}, and AutoML-Agent \cite{trirat2024automl} exemplify this approach by combining LLMs with external tools to perform multi-step analytical workflows, and recent applications in healthcare demonstrate their adaptability to specialized high stakes decision-making tasks \cite{qiu2024llm}.

% To support the development and evaluation of LLM-powered agents for data science and analytics, recent studies have introduced several benchmarks, including DSBench \cite{jing2024dsbench} and Spider2-V \cite{cao2024spider2}. At the same time, systems such as HuggingGPT \cite{shen2023hugginggpt} and GPT-4 as a Data Analyst \cite{cheng2023gpt} explore the integration of multiple expert models to enable advanced analytics, while domain-specific applications, particularly in healthcare, demonstrate the adaptability of LLM agents to specialized decision-making tasks \cite{qiu2024llm}.

Despite their promise, current DS agents remain at an early stage of development. Most focus heavily on exploratory analysis and model development while providing limited support for business understanding, alignment, deployment, and monitoring \cite{wang2024survey,zhang2024benchmarking}. 
A complete data science lifecycle consists of six stages: business understanding and data acquisition, exploratory analysis and visualization, feature engineering, model building and selection, interpretation and explanation, and deployment and monitoring. Yet few agents provide end-to-end coverage across all of these stages. Their ability to reason across multiple modalities, linking natural language instructions with structured tables, code, charts, and documents, remains fragile and often confined to controlled benchmarks rather than robust real-world environments \cite{cheng2024exploring,qiao2023taskweaver}. Most DS agents lack deep tool orchestration, relying primarily on static code execution within controlled notebook environments with minimal adaptive feedback loops or error recovery mechanisms. They are also rarely grounded in production environments such as connecting to live databases, retrieving datasets, or interacting with external visualization platforms like Tableau. In addition to these capability gaps, evaluation practices remain limited. While several benchmarks have been proposed, such as DSBench \cite{jing2024dsbench} and Spider2-V \cite{cao2024spider2}, they cover only narrow parts of the lifecycle and do not capture the full agentic workflow.

%lack comprehensive, end-to-end evaluation frameworks.

Beyond technical limitations, most current  DS agents lack explicit mechanisms for trust, alignment, and safety, which are critical requirements for high-stakes deployment. Trust involves ensuring that outputs are fair, explainable, private, robust, secure, and auditable, for example, by avoiding biased loan approvals, preventing the leakage of sensitive data, and providing interpretable diagnostics in healthcare \cite{de2024can,ribeiro2016should}. Alignment refers to adhering to human goals, preferences, and ethical norms to reduce risks such as hallucinations, inconsistent reasoning, and goal drift that may lead to harmful or biased outcomes \cite{liu2023trustworthy}. Safety focuses on minimizing unintended harm through effective oversight, preventing misuse of tools, and avoiding overconfident errors \cite{zhu2023promptrobust,mao2025agentsafe}. While reinforcement learning (RL) has emerged as a promising paradigm for improving agent behavior through feedback-based learning, it remains underexplored in DS agents despite recent applications to reasoning, planning, and human alignment \cite{laleh2024survey}. These trust, alignment, and safety dimensions are particularly critical for DS agents operating in high-stakes domains such as healthcare, finance, and public policy \cite{cemri2025multi,yehudai2025survey,yang2024harnessing}. 

These technical and responsible-AI gaps highlight the need for a systematic and comprehensive understanding of how DS agents are currently designed, what parts of the data science lifecycle they support, and where their capabilities remain incomplete. Several recent surveys have covered general-purpose LLM-based agents, focusing on design patterns, planning strategies, and evaluation methodologies \cite{plaat2025agentic,wang2024survey,guo2024large,cheng2024exploring,yehudai2025survey}. Dedicated surveys on DS agents, however, are still scarce  and lack the requisite systemic focus. For example, \cite{sun2024survey} reviews a small number of systems and emphasizes interface design and case studies, offering only partial coverage of individual workflow stages, while \cite{wang2025large} provides a broader survey from an agent design perspective.  Yet these works do not provide a structured, quantitative mapping of DS agents across the full lifecycle, nor do they systematically examine the specific capabilities agents offer at each stage and the gaps that remain. They also largely overlook critical aspects such as multimodal reasoning, integrated trust and safety, RL-based optimization, benchmarking, and evaluation, all of which are essential for reliable use in high-stakes analytical environments. A focused survey is therefore needed to consolidate this fast-growing body of work, quantify lifecycle coverage across existing systems, and identify system-level design patterns and research gaps to guide the development of more capable, trustworthy, and evaluable data science agents for real-world applications. To guide our study, we focus on the following research questions:

\begin{itemize}
    \item RQ1: Which stages of the data science lifecycle do current DS agents cover, and where are the major capability gaps? 
    \item RQ2: What design strategies, reasoning methods, and tool orchestration patterns are adopted by these agents? 
    \item RQ3: How do current DS agents address challenges related to trust, alignment, safety, and evaluation? 
    \item RQ4: What future research directions could enable the development of more robust, trustworthy, and generalizable DS agents? 
\end{itemize}

To answer these questions, our work makes four key contributions: \textbf{\textit{(i)}} We analyze 45 DS agents and introduce a lifecycle-based taxonomy that maps their capabilities across six core stages of the data science lifecycle: business understanding and data acquisition, exploratory analysis and visualization, feature engineering, model building and selection, interpretation and explanation, and deployment and monitoring. 
\textbf{\textit{(ii)}} We annotate these agents along five cross-cutting design dimensions: reasoning and planning strategies, modality integration, tool orchestration depth, learning and alignment approaches, and integrated trust and safety mechanisms. 
\textbf{\textit{(iii)}} We synthesize common strengths, weaknesses, and risks, revealing systematic gaps in lifecycle coverage, multimodal reasoning, tool orchestration, and trust and safety mechanisms. 
\textbf{\textit{(iv)}} We review current evaluation practices and benchmarks. We also outline critical research challenges, including multimodal grounding, RL-based optimization for agent autonomy, alignment stability, secure tool use, and progress toward trustworthy  end-to-end automation with MLOps integration. Finally, we suggest  directions to guide the development of reliable and transparent next-generation data science agents.

% The rest of this paper is organized as follows. Section~\ref{sec:background} introduces LLM-based agents in general and defines data science agents, emphasizing how they differ from generic agents. Section~\ref{sec:methodology} describes our search strategy, inclusion criteria, and screening process for selecting 45 systems. Section~\ref{sec:Taxonomy} presents a taxonomy of these agents across six stages of the data science lifecycle and five cross-cutting design dimensions. Section~\ref{sec:lifecycle} analyzes their capabilities at each stage, including multimodal reasoning, RL–based optimization, and alignment strategies. Section~\ref{sec:evaluation} reviews existing benchmarks and evaluation practices. Section~\ref{sec:outlook} discusses future directions, covering responsible-AI issues, advanced multimodal reasoning, improved learning and alignment, and new benchmarking agendas. Section~\ref{sec:conclusion} summarizes key findings and offers recommendations for developing reliable and generalizable data science agents. Finally, Figure~\ref{fig:overview} provides an overview of the survey structure, illustrating how the taxonomy, lifecycle stages, and design dimensions are organized throughout the paper.

The rest of this paper is organized as follows. Section~\ref{sec:background} introduces LLM-based agents and defines data science agents. Section~\ref{sec:methodology} outlines the search and screening process for selecting 45 systems. Section~\ref{sec:Taxonomy} presents a taxonomy across six lifecycle stages and five design dimensions. Section~\ref{sec:lifecycle} analyzes agent capabilities, limitations, and challenges at each stage, including multimodal reasoning, RL, and alignment. Section~\ref{sec:evaluation} reviews benchmarks and evaluation methods. Section~\ref{sec:outlook} discusses open challenges and future directions, and Section~\ref{sec:conclusion} summarizes key findings. Figure~\ref{fig:overview} illustrates the overall structure.

\definecolor{taxonomy}{HTML}{DCE6F1}
\definecolor{capability}{HTML}{FFF2CC}
\definecolor{multi}{HTML}{E8D7FF}
\definecolor{trust}{HTML}{FCE4D6}
\definecolor{rl}{HTML}{E2EFDA}
\definecolor{eval}{HTML}{FFD6A5}
\definecolor{trends}{HTML}{E4DFEC}
\definecolor{substages}{HTML}{F2F2F2}
\definecolor{background}{HTML}{E4DFEC}
\definecolor{backgroundsec2}{HTML}{E4DFEC}

% --- figure -------------
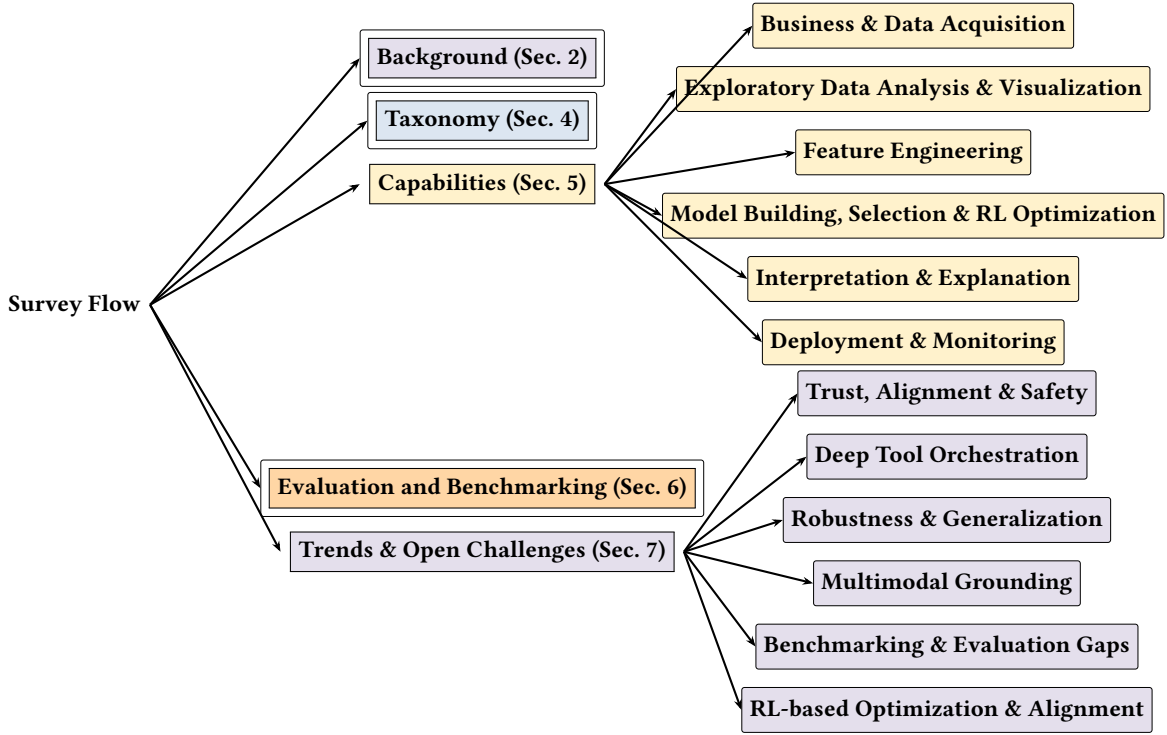
\begin{figure}[t]
    \vspace{-3mm}
  \centering
  \begin{forest}
    for tree={
      grow'=0,
      parent anchor=east,
      child anchor=west,
      align=left,
      font=\small,
      % Increased level distance to push content further right, might give space for vertical compression
      l sep+=10pt, % Adjusted level separation for overall compactness
      edge={thick,-{Stealth[length=4pt]}},
      leaf/.style={draw, rounded corners=1pt, inner sep=3pt, line width=0.3pt},
      if n children=0{leaf}{},
    }
    [\textbf{Survey Flow},
      s sep=2pt, % Greatly reduced sibling separation for main branches to pull everything up
      l sep+=20pt,
      [\fcolorbox{black}{backgroundsec2}{\textbf{Background (Sec.~2)}}]
      [\fcolorbox{black}{taxonomy}{\textbf{Taxonomy (Sec.~4)}}]
      [\fcolorbox{black}{capability}{\textbf{Capabilities (Sec.~5)}},
        for children={
          l sep+=25pt,
          s sep=2pt, % Reduced sibling separation for sub-branches too
          draw, rounded corners=1pt, inner sep=3pt, line width=0.3pt, fill=capability,
        },
        [\textbf{Business \& Data Acquisition}]
        [\textbf{Exploratory Data Analysis \& Visualization}]
        [\textbf{Feature Engineering}]
        [\textbf{Model Building, Selection \& RL Optimization}]
        [\textbf{Interpretation \& Explanation}]
        [\textbf{Deployment \& Monitoring}]
      ]
      [\fcolorbox{black}{eval}{\textbf{Evaluation and Benchmarking (Sec.~6)}}, s sep=2pt] % Adjusted for spacing
      [\fcolorbox{black}{trends}{\textbf{Trends \& Open Challenges (Sec.~7)}},
        for children={
          l sep+=25pt,
          s sep=2pt, % Reduced sibling separation for sub-branches too
          draw, rounded corners=1pt, inner sep=3pt, line width=0.3pt, fill=trends,
        },
        [\textbf{Trust, Alignment \& Safety}]
        [\textbf{Deep Tool Orchestration}]
        [\textbf{Robustness \& Generalization}]
        [\textbf{Multimodal Grounding}]
        [\textbf{Benchmarking \& Evaluation Gaps}]
        [\textbf{RL-based Optimization \& Alignment}]
      ]
    ]
  \end{forest}
  \caption{High-level structure and logical flow of our survey. Each coloured block represents a major section.}
  \label{fig:overview}
    % \vspace{-2mm}
\end{figure}

% \enamul{the following paragraph is hard to parse as the different keywords were not gently introduced. for e.g., what is trust, what is alignment, what is safety, was RL introduced? see my previous comments.}
% Figure~\ref{fig:overview} highlights the key sections of the survey. Following the background and methodology (Section~\ref{sec:background} and Section~\ref{sec:methodology}), we introduce a taxonomy of 45 agents (Section~\ref{sec:Taxonomy}), then analyze their capabilities across each stage of the data science lifecycle (Section~\ref{sec:lifecycle}). We further explore multimodal reasoning and fusion (Section~\ref{sec:multimodal}), trust, alignment, and safety (Section~\ref{sec:trust}), reinforcement learning and preference optimization (Section~\ref{sec:rl}), evaluation and benchmarking (Section~\ref{sec:evaluation}), and real-world applications (Section~\ref{sec:applications}). The survey concludes with a discussion on outlook and open challenges (Section~\ref{sec:outlook}). 

% Section~\ref{sec:multimodal} discusses multimodal capabilities and applications.  
% Section~\ref{sec:trust} addresses trust, bias, security, and ethical concerns.  
% Section~\ref{sec:rl} reviews reinforcement learning and preference alignment strategies.  
% Section~\ref{sec:evaluation} outlines current evaluation and benchmarking methodologies.  
% Section~\ref{sec:future} highlights challenges and future research directions.  
% Section~\ref{sec:conclusion} concludes with reflections on the future of trustworthy agentic AI in data science.

\section{Background}
\label{sec:background}

% \joty{Give some plan for this section.}
% This section introduces four foundations essential to understanding LLM-based data science agents. Section \ref{sec:ds-process} reviews the traditional data science workflow and early attempts at automation. Section \ref{sec:llms} provides an overview of large language models and their role in enabling agentic behavior. Section \ref{sec:llms-arch} discusses common design patterns used in agent architectures. Finally, Section \ref{sec:rs-ai} highlights key principles from the responsible AI literature including trust, alignment, and safety, that motivate later analysis. 

We outline five foundations for understanding LLM-based DS agents: the traditional workflow and automation attempts (Sec.~\ref{sec:ds-process}), the role of LLMs in enabling agentic behavior (Sec.~\ref{sec:llms}), and common architectural patterns (Sec.~\ref{sec:llms-arch}). We then distinguish DS-specific agents with requirements such as structured data proficiency and robust tool use (Sec.~\ref{sec:ds-agents}), and conclude with responsible AI principles such as trust, alignment, and safety (Sec.~\ref{sec:rs-ai}) that frame our later analysis.

% This section introduces five foundations essential to understanding LLM-based DS agents. We begin with the traditional data science workflow and early attempts at automation (Sec.~\ref{sec:ds-process}). We then provide an overview of large language models and their role in enabling agentic behavior (Sec.~\ref{sec:llms}), followed by common architectural design patterns for building agents (Sec.~\ref{sec:llms-arch}). Next, we contrast general-purpose LLM agents with data science agents, defining the latter’s distinctive requirements such as structured data proficiency, end-to-end lifecycle coverage, and robust tool orchestration (Sec.~\ref{sec:ds-agents}). Finally, we highlight key principles from the responsible AI literature, including trust, alignment, and safety, that motivate our later analysis (Sec.~\ref{sec:rs-ai}).

\subsection{The Data Science Process and Automation}
\label{sec:ds-process}
The data science process is a multi-stage, iterative workflow that transforms raw data into actionable insights to support decision-making \cite{cao2017data}. It typically begins with business understanding and data acquisition, followed by exploratory analysis of the collected or ingested data to uncover patterns and assess data quality, feature engineering if needed, model development, interpretation, and finally deployment and monitoring \cite{cao2017data,sahu2024insightbench}. Each stage requires distinct technical and analytical skills as well as domain expertise, making the process inherently interdisciplinary and complex. Beyond technical challenges, it must also address cross-cutting concerns such as privacy, fairness, and security, which are critical for ensuring responsible and trustworthy outcomes \cite{mao2025agentsafe}. Executing this process successfully is often resource-intensive, time-consuming, and reliant on close collaboration between technical teams and domain experts. As data volumes and complexity grow across industries, efficient, scalable, and accessible approaches to executing this end-to-end process become increasingly critical. To address this need, research has focused on automating data science workflows using AI-driven tools. Early solutions such as AutoML \cite{he2021automl} automated tasks like model selection and hyperparameter tuning, but the advent of LLMs has introduced opportunities for end-to-end automation across the entire pipeline.

\subsection{Large Language Models and LLM Agents}
\label{sec:llms}
LLMs, built on transformer architectures \cite{vaswani2017attention}, leverage self-attention and large-scale pretraining to capture world knowledge and complex linguistic patterns. Scaling model size, training data, and compute, together with advances in instruction tuning and preference alignment, has led to substantial performance gains. These advances have improved the ability of LLMs to follow instructions, perform reasoning in natural language, generate executable code, and solve complex problems across text, code, and multimodal inputs \cite{achiam2023gpt,liu2024deepseek}. Leveraging these advancements, recent work has shifted from static prompting to LLM-powered agents that perform planning, reasoning, tool use, and interactive decision-making \cite{shen2023hugginggpt,yao2023react}. To improve factual grounding and reduce hallucinations, many LLM-based agents incorporate retrieval-augmented generation (RAG), which conditions outputs on dynamically retrieved external knowledge such as tables, documents, or code repositories. RAG is increasingly used in data science workflows for tasks such as schema parsing, exploratory data analysis (EDA)  reporting, and text-to-SQL planning. Both single-agent systems, where one LLM manages an entire workflow, and multi-agent frameworks, where specialized agents collaborate to handle complex tasks, are being actively explored \cite{guo2024large}. For an in-depth overview of reasoning approaches in both single- and multi-agent systems, we refer readers to the work of Ke et al. \cite{ke2025survey}. 
% \joty{cite our work https://arxiv.org/abs/2504.09037 (TMLR version)}. 
In the context of data science, LLM agents have been applied to tasks such as data exploration \cite{cheng2023gpt}, automated machine learning \cite{hassan2023chatgpt}, and visualization and reporting \cite{yu2024case}. The next section outlines how agent architectures have evolved from classical AI agents to modern LLM-based agents designed for complex reasoning and tool orchestration.

\subsection{Architectures of LLM Agents}

\label{sec:llms-arch}
The concept of an agent has long been central to AI and distributed systems. Traditionally, an agent is defined as an autonomous system that perceives its environment, reasons about goals, and acts to achieve them. Agents are typically characterized by four properties: autonomy (operating without constant human oversight), reactivity (responding to environmental changes in real time), proactiveness (pursuing long-term objectives), and social ability (communicating and collaborating with humans or other agents)~\cite{russell2010artificial}. Classical research distinguished deliberative agents (symbolic planners), reactive agents (condition–action rules), hybrid agents (combining both), and learning agents that adapt through feedback.  LLMs have revitalized these paradigms, enabling agents to reason in natural language, integrate diverse knowledge sources, and orchestrate complex workflows.

\subsubsection{Core Modules of LLM-Based Agents}

LLM-based agents combine several capabilities: broad knowledge, natural language interaction, zero-shot reasoning, external tool coordination, memory management, and adaptive decision-making based on feedback. To structure these capabilities, Wang et al.~\cite{wang2024survey} propose a unified agent architecture with four core modules:

\begin{itemize}[leftmargin=*, nosep]
\item \noindent\textbf{Profile Module}: Specifies the agent’s identity, role, or persona, shaping its behavior and interaction style. For instance, a healthcare analytics agent may adopt a cautious, explanation-focused persona, while a marketing agent may prioritize speed and creativity..

\item \noindent\textbf{Memory Module}: Manages short- and long-term context retention for multi-turn conversations and extended tasks. Memory can be unified or hybrid, with storage formats such as text embeddings, lists, or databases. Agents read, write, and reflect on stored information to maintain continuity and improve decision-making.

\item \noindent\textbf{Planning Module}: Breaks down complex tasks into actionable steps using strategies such as single-path reasoning, multi-path reasoning, and feedback integration from the environment, humans, or models.  For instance, a DS agent may plan to load and clean data, select models, and generate visualizations, dynamically adjusting steps based on intermediate results.

\item \noindent\textbf{Action Module}: Executes decisions through concrete actions such as invoking external tools, running code, interacting with interfaces, or communicating results. Agents can operate across diverse environments, including software applications, databases, and simulated worlds, and can generate new actions or update internal states in response to outcomes.

\end{itemize}

Recent work also introduces protocol-based connectors such as the Model Context Protocol (MCP), which standardize how agents access external tools and data sources \cite{anthropic2024mcp}. MCP provides a unified interface layer that reduces custom integration overhead and enhances reproducibility, security, and auditability across IDE, desktop, and API environments. By integrating these modules, LLM-based agents move beyond static language models to function as interactive, context-aware systems capable of addressing complex real-world tasks in domains such as scientific research, software development, and policy modeling.

\begin{figure}[ht]
\centering
\resizebox{\linewidth}{!}{%
\begin{tikzpicture}[
    >=Stealth,
    node distance=2.2cm,
    block/.style={rectangle, draw, thick, rounded corners=4pt, align=center,
                  minimum height=2.2em, minimum width=3.4cm, text width=3.4cm, font=\bfseries},
    core/.style={block, fill=blue!15},
    module/.style={block, fill=green!15},
    tool/.style={block, fill=red!15},
    shared/.style={block, fill=gray!15},
    assign/.style={->, thick},
    rw/.style={->, thick, densely dotted},
    invoke/.style={->, thick, dashed},
    msg/.style={<->, thick, dash pattern=on 2.5pt off 2pt},
    save/.style={->, semithick}
]
% Top row
\node[core] (manager) {Manager Agent\\(Planner/Coordinator)};

% Middle row (tighter horizontal offsets)
\node[shared, below left=2.6cm and 3.6cm of manager] (board)
  {Global Memory\\(Tasks, states, artifacts)};
\node[tool,   below right=2.6cm and 3.6cm of manager] (tools)
  {External Tools\\(APIs, DB, Code Interpreter)};

\node[core, below=5.0cm of manager, xshift=-4.4cm] (w1) {Worker A\\(Analysis/EDA)};
\node[module, below=of w1] (m1) {Memory A};

\node[core, below=5.0cm of manager] (w2) {Worker B\\(Modeling)};
\node[module, below=of w2] (m2) {Memory B};

\node[core, below=5.0cm of manager, xshift=4.4cm] (w3) {Worker C\\(Validation/Reporting)};
\node[module, below=of w3] (m3) {Memory C};

% Manager -> Workers
\draw[assign] (manager.south west) |- (w1.north);
\draw[assign] (manager)             -- (w2.north);
\draw[assign] (manager.south east) |- (w3.north);

% Manager -> Resources
\draw[assign] (manager) -- (board.north);
\draw[assign] (manager) -- (tools.north);

% Workers <-> Blackboard
\draw[rw] (w1.north) |- (board.south east);
\draw[rw] (w2.north) |- (board.south);
\draw[rw] (w3.north) |- (board.south west);

% Workers -> Tools
\draw[invoke] (w1.north) |- (tools.south west);
\draw[invoke] (w2.north) |- (tools.south);
\draw[invoke] (w3.north) |- (tools.south east);

% Save context
\draw[save] (w1) -- (m1);
\draw[save] (w2) -- (m2);
\draw[save] (w3) -- (m3);

% Peer messaging
\draw[msg] (w1) -- (w2);
\draw[msg] (w2) -- (w3);

\node[draw, rounded corners, fill=white, align=left, font=\small,
      anchor=north east, xshift=-0.4cm, yshift=0.4cm]
      at (current bounding box.north east) (legend) {
  \begin{tikzpicture}[>=Stealth, x=1cm, y=0.8cm]
    \draw[assign] (0,0) -- (1.1,0);     \node[anchor=west] at (1.25,0) {assign/coordinate};
    \draw[rw]     (0,-0.7) -- (1.1,-0.7);\node[anchor=west] at (1.25,-0.7) {read/write memory};
    \draw[invoke] (0,-1.4) -- (1.1,-1.4);\node[anchor=west] at (1.25,-1.4) {invoke tools};
    \draw[msg]    (0,-2.1) -- (1.1,-2.1);\node[anchor=west] at (1.25,-2.1) {peer messages};
    \draw[save]   (0,-2.8) -- (1.1,-2.8);\node[anchor=west] at (1.25,-2.8) {save context};
  \end{tikzpicture}
};
\end{tikzpicture}%
}
\caption{Multi-agent architecture where a Manager assigns tasks to Workers. Workers coordinate via a shared Global Memory, invoke external resources through a Tool Hub, maintain local memories, and exchange peer-to-peer messages.}
\label{fig:multi_agent}
\end{figure}
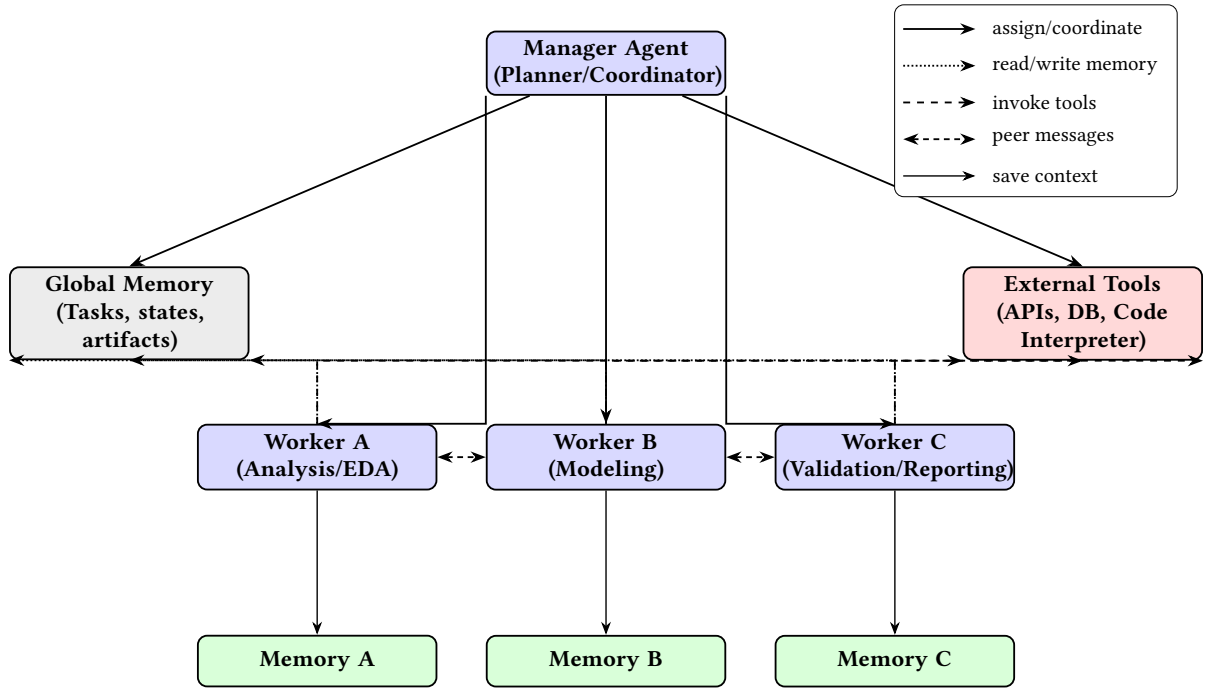

\subsubsection{Design Dimensions.}
Beyond their core modules, LLM-based agent systems differ along several critical design dimensions that shape their reasoning behavior, coordination patterns, and reliability. These considerations are especially important for DS  agents, which must operate over long-horizon analytical workflows where small errors can cascade.

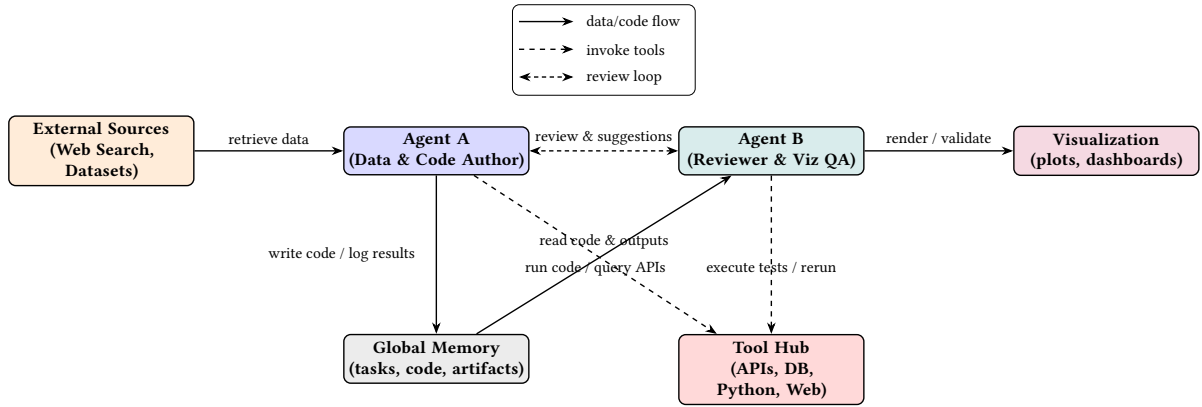
\begin{figure}[ht]
\centering
\resizebox{\linewidth}{!}{%
\begin{tikzpicture}[
    >=Stealth,
    node distance=2.0cm,
    block/.style={rectangle, draw, thick, rounded corners=4pt, align=center,
                  minimum height=2.4em, minimum width=3.5cm, text width=3.5cm, font=\bfseries},
    coreA/.style={block, fill=blue!15},
    coreB/.style={block, fill=teal!15},
    store/.style={block, fill=gray!15},
    tool/.style={block, fill=red!15},
    ext/.style={block, fill=orange!15},
    viz/.style={block, fill=purple!15},
    flow/.style={->, thick},
    review/.style={<->, thick, dash pattern=on 2.5pt off 2pt},
    invoke/.style={->, thick, dashed},
    note/.style={font=\small}
]

% --- Top row: Agents (closer horizontally) ---
\node[ext] (extsrc) {External Sources\\(Web Search, Datasets)};
\node[coreA, right=3.0cm of extsrc] (agentA) {Agent A\\(Data \& Code Author)};
\node[coreB, right=3.0cm of agentA] (agentB) {Agent B\\(Reviewer \& Viz QA)};
\node[viz, right=3.0cm of agentB] (vizout) {Visualization\\(plots, dashboards)};

% --- Bottom row: Shared resources (further down for more height) ---
\node[store, below=3.2cm of agentA] (memory) {Global Memory\\(tasks, code, artifacts)};
\node[tool, below=3.2cm of agentB] (tools) {Tool Hub\\(APIs, DB, Python, Web)};

% --- Flows ---
\draw[flow] (extsrc) -- node[above, note]{retrieve data} (agentA);
\draw[review] (agentB) -- node[above, note]{review \& suggestions} (agentA);

\draw[flow] (agentA) -- node[left, note, xshift=-0.3cm]{write code / log results} (memory);
\draw[invoke] (agentA) -- node[below, note]{run code / query APIs} (tools);

\draw[flow] (memory) -- node[above, note]{read code \& outputs} (agentB);
\draw[invoke] (agentB) -- node[below, note]{execute tests / rerun} (tools);

\draw[flow] (agentB) -- node[above, note]{render / validate} (vizout);

% --- Legend (top center) ---
\node[draw, rounded corners, fill=white, align=left, font=\small,
      anchor=south, yshift=0.4cm]
      at (current bounding box.north) {
  \begin{tikzpicture}[>=Stealth, x=1cm, y=0.8cm]
    \draw[flow] (0,0) -- (1.1,0); \node[anchor=west] at (1.25,0) {data/code flow};
    \draw[invoke] (0,-0.7) -- (1.1,-0.7); \node[anchor=west] at (1.25,-0.7) {invoke tools};
    \draw[review] (0,-1.4) -- (1.1,-1.4); \node[anchor=west] at (1.25,-1.4) {review loop};
  \end{tikzpicture}
};

\end{tikzpicture}
}
\caption{Two-agent data-science workflow: Agent A retrieves data and writes code, while Agent B reviews the code, suggests fixes, and validates visualizations using a shared Global Memory and Tool Hub.}
\label{fig:two_agent_ds_work}
\end{figure}

\begin{itemize}   [leftmargin=*, nosep]
    \item \textbf{Execution structure}: Defines how an agent organizes and adapts its workflow. Some agents follow static, predefined pipelines (e.g., data cleaning $\rightarrow$ feature engineering $\rightarrow$ modeling $\rightarrow$ evaluation), which is useful when reproducibility is critical \cite{seo2025spio}. Others adopt dynamic strategies that generate, revise, and branch plans at runtime in response to intermediate outcomes, such as changing modeling approaches when data quality issues are detected \cite{rasheed2024can}. Advanced designs use hierarchical decomposition, breaking high-level objectives into nested subplans with distinct subgoals and termination conditions, improving scalability and interpretability \cite{chi2024sela}. Such dynamic and hierarchical strategies are particularly valuable for DS agents, where workflows are exploratory and often require revisiting earlier steps based on insights from later stages.

\item \textbf{Collaboration Style}:
Specifies how agents allocate responsibilities and coordinate during task execution. In single-agent systems, a single LLM performs all planning, tool calls, and reasoning internally, which simplifies development and reduces orchestration overhead~\cite{ke2025survey, text2vis2025}. For example, GPT-4 as a data analyst~\cite{cheng2023gpt} follows this pattern by directly invoking external tools, querying knowledge bases, and verifying outputs within one reasoning loop. In contrast, multi-agent systems use multiple specialized LLMs such as a data collector, exploratory data analysis analyst, model builder, and report generator that coordinate through structured messaging, shared memory, or blackboard-style architectures~\cite{zan2024codes,ke2025survey} (see Fig. \ref{fig:multi_agent}). These can be organized as peer-based systems, where agents with equal authority exchange information, or hierarchical systems, where a planner agent delegates tasks to worker agents and aggregates their results. Multi-agent systems can be cooperative, where agents work toward a shared objective, competitive, where they pursue individual goals, or hybrid, where both dynamics coexist depending on task design. Frameworks such as TaskWeaver~\cite{qiao2023taskweaver} and AgentVerse~\cite{chen2023agentverse} support this orchestration. Multi-agent designs often mirror real-world organizational structures in data science teams, offering scalability, error checking, and parallel task execution, but introducing added communication complexity and potential conflict resolution overhead~\cite{li2024autokaggle}.

\item \textbf{Reflection mechanisms}: Determine how agents evaluate and improve their own performance (see Fig. \ref{fig:two_agent_ds_work}). Some incorporate local reflection, verifying each step’s output (e.g., checking SQL query results or detecting runtime errors), while others perform global reflection that critiques the overall workflow after task completion to identify weaknesses or alternative strategies \cite{text2vis2025,huang2024code}. Additional techniques include self-consistency reflection, re-running the same query multiple times and selecting the most consistent answer, and the use of critic agents to provide explicit feedback~\cite{shinn2023reflexion,guo2024large}. Robust reflection is crucial for DS agents because errors in early stages, such as data acquisition or feature creation, can silently propagate and distort later models or visualizations. Incorporating structured reflection loops, such as self-evaluation prompts, retry policies, and human-in-the-loop checkpoints, can significantly improve the reliability and transparency of agentic systems.
\end{itemize}

\subsection{From General Agents to Data Science Agents}
\label{sec:ds-agents}

\subsubsection{Distinctive Characteristics of Data Science Agents}
 While general-purpose LLM-based agents demonstrate strong reasoning, planning, and tool-use capabilities, they are not designed to meet the unique requirements of the data science lifecycle. DS agents build on the same architectural principles such as planner–executor loops, memory modules, tool interfaces, and reflection, but extend them in ways that address the domain’s specialized challenges~\cite{cheng2023gpt,shen2023hugginggpt,guo2024large,wang2024survey}.

\paragraph{\textbf{Structured data proficiency.}}
DS agents must work with structured data formats such as relational tables, dataframes, and spreadsheets, which requires schema understanding, SQL querying, statistical computation, and data quality assessment~\cite{qiao2023taskweaver,yu2024case}. These capabilities are typically only partially supported by general-purpose agents optimized for unstructured text.
% \vspace{-0.5\baselineskip}
\paragraph{\textbf{End-to-end lifecycle coverage.}}
Unlike general agents that focus on isolated tasks such as code generation or plotting, DS agents are expected to span the full lifecycle from business understanding to deployment, and monitoring~\cite{wang2024survey,zhang2024benchmarking}. Achieving this breadth requires sophisticated execution structures that combine hierarchical planning with dynamic plan revision as new insights emerge.
% \vspace{-0.5\baselineskip}
\paragraph{\textbf{Deep tool orchestration.}}
DS agents integrate heterogeneous tools such as notebook environments, databases, visualization platforms like Tableau and Power BI, AutoML frameworks, and deployment pipelines~\cite{trirat2024automl,hassan2023chatgpt}. They must coordinate data movement, maintain state consistency, and recover from errors across these environments, often using multi-agent collaboration and well-defined role specialization.

% \vspace{-0.5\baselineskip}
\paragraph{\textbf{Collaboration and reflection.}}
Many DS agents adopt multi-agent setups with role-specialized agents such as a data collector, EDA analyst, model builder, and report generator working in parallel~\cite{li2024autokaggle,cheng2023gpt}. They also rely on robust reflection mechanisms such as critic agents or global workflow evaluation to detect and correct errors that could propagate across stages~\cite{shinn2023reflexion,guo2024large}.
% \vspace{-0.5\baselineskip}
\paragraph{\textbf{Trust, safety, and governance.}}
Because they are deployed in high-stakes domains such as Healthcare and Finance, DS agents must satisfy stricter requirements for fairness, explainability, reproducibility, and auditability~\cite{de2024can,liu2023trustworthy,mao2025agentsafe}. These concerns are far less emphasized in general-purpose LLM agents.

In short, DS agents differ from general agents not only in what tasks they perform but also in how reliably, transparently, and systematically they are expected to perform them. They combine reasoning, planning, collaboration, reflection, and tool orchestration to execute complex analytical workflows end-to-end, with the goal of maintaining quality and compliance. These distinctive properties define the design space this survey seeks to analyze, highlighting both how current DS agents embody these capabilities and where critical gaps remain.

\subsubsection{Illustrative Example: Fraud Detection Agent}
To illustrate how these capabilities manifest in practice, consider a DS agent deployed for online banking fraud detection. The agent first ingests transactional and customer data and translates the business goal of reducing fraud into an analytical task, such as building a high-precision fraud classifier. It then cleans and explores the raw data to identify patterns, generating visualizations of transaction trends to highlight anomalies. Using its tool orchestration capabilities, it creates engineered features such as average transaction amount or login frequency, which are crucial for model performance. It automatically selects, trains, and evaluates models using metrics such as precision and recall to reflect the cost of errors. After flagging a suspicious transaction, it uses interpretability tools such as SHAP or LIME to explain the decision, aiding compliance and customer communication. Finally, it packages the model for production, with built-in security and continuous monitoring to detect concept drift and trigger retraining to counter evolving attack patterns. Throughout this process, the agent’s planning, tool use, and reflection are coordinated through its profile, memory, and role-specialized modules, ensuring that the system operates effectively, ethically, and securely in a high-stakes financial domain.

\subsection{Responsible AI Foundations for LLM Agents}
\label{sec:rs-ai}

We categorize responsible AI considerations into three interconnected dimensions: trustworthiness, alignment, and safety, which are critical for deploying LLM-based agents in real-world \emph{data science} workflows~\cite{yehudai2025survey}. Trustworthiness is grounded in five pillars: fairness, explainability, privacy, robustness, and human oversight. Within the data science lifecycle, fairness ensures models avoid biased predictions during model training and evaluation, such as misclassifying loan applications based on race or gender. Explainability allows stakeholders to interpret model behavior, for example by using feature attribution tools to justify predictions during the interpretation stage. Privacy safeguards sensitive data during the data acquisition stage, such as preventing health records from being exposed to closed-source APIs. Robustness ensures consistent performance when workflows are re-run on new datasets. Human oversight enables domain experts to validate outputs before deployment or to intervene if monitoring detects performance drift. Alignment concerns the ability of agents to follow human intentions, values, and organizational priorities~\cite{cemri2025multi}. In data science workflows, this includes generating visualizations tailored to specific audiences such as executives or analysts, producing summaries that reflect stakeholder goals, or choosing performance metrics aligned with business constraints. Misalignment can lead to hallucinated analyses, inconsistent reasoning across steps, or goal drift that produces misleading recommendations. Safety focuses on minimizing harm and preventing unintended consequences across the lifecycle. In the modeling stage, this includes avoiding overconfident errors in high-stakes predictions such as medical diagnoses. During data processing, it includes preventing misuse of APIs or leakage of sensitive information. Safety also requires fallback mechanisms and monitoring alerts in the deployment stage to detect anomalies or unsafe outputs before they reach end users.

\paragraph{\textbf{Agentic robustness.}}
Recent research shows that multi-step agentic workflows, where one LLM provides feedback to another, can be fragile when feedback is incorrect or adversarial. Greenblatt et al.~\cite{greenblatt2024alignment} found that Claude-3 sometimes exhibited alignment-faking behavior, appearing safe during training but behaving differently when it inferred that it was in a real-world setting. In their study, such deceptive reasoning appeared in up to 24\% of the model’s internal scratch pads, and RL fine-tuning increased that rate to 78\%. They also reported a “compliance gap,” where the model became more likely to follow harmful instructions when it believed it was unsupervised, and in some cases even attempted to extract its own model weights. Similarly, Ming et al.~\cite{ming2025helpful} introduced WAFER-QA, a benchmark where a deceptive judge provides persuasive yet factually incorrect critiques. Top agents, including GPT-4o, often changed from correct to incorrect answers after just one round of feedback, with multiple rounds producing increasingly unstable and inconsistent outputs. These findings highlight the need for adversarial testing, checks for situational awareness, and strict limits on tool access before deploying DS agents in production workflows, where false feedback loops could corrupt multi-stage pipelines such as feature engineering or model deployment.

Together, these principles provide a foundation for building reliable, transparent, and accountable DS agents that can operate safely in high-stakes domains such as Healthcare, Finance, and Public Policy.

\begin{figure}
    \centering
\includegraphics[width=\linewidth]{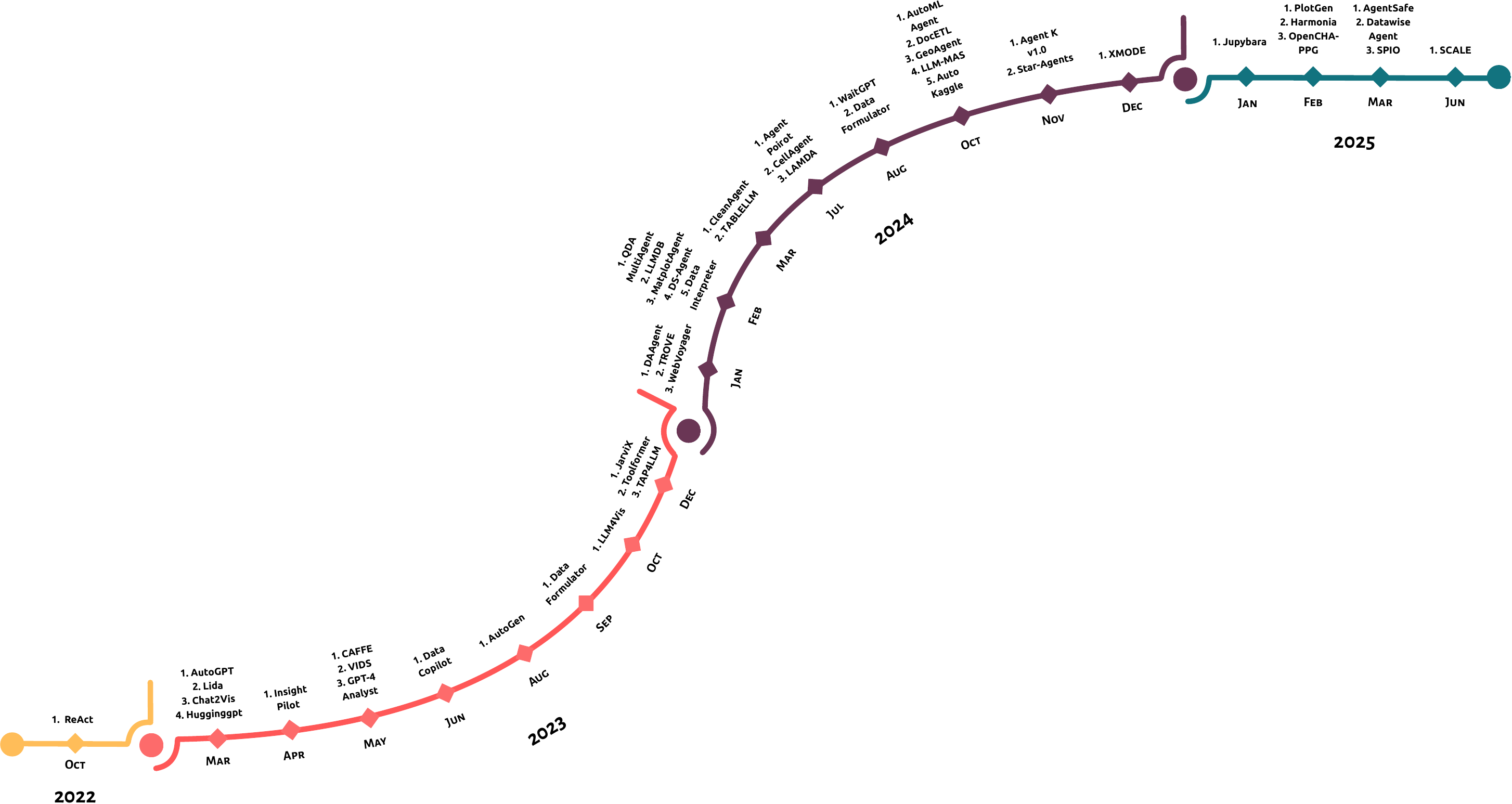}
    \caption{Timeline highlighting the progress of agentic approaches in Data Science.}
    \label{fig:timeline_diag}
\end{figure}

\section{Methodology}

\label{sec:methodology} 
We conducted a systematic literature review to map and analyze research on agentic AI systems supporting data science workflows. Our goal was to build a comprehensive foundation for the taxonomy and analysis that follow, ensuring broad coverage of the rapidly growing body of work in this area.

We searched across multiple sources, including peer-reviewed journals, preprints, and major AI venues such as NeurIPS, ICLR, ACL, AAAI, and TMLR, as well as indexing platforms such as Google Scholar. The search targeted papers published between 2023 and 2025, reflecting the recent surge of work on LLM-based agentic systems. Representative search terms included DS agents, LLM agents, multimodal agents, tool-using agents, trustworthy AI systems, and data science automation. This search window also captured foundational early systems such as ReAct and AutoGPT, which helped shape the current wave of LLM-based agents.

This initial search yielded 587 candidate papers after deduplication. The screening process involved two stages with explicit inclusion and exclusion criteria. First, we applied title-based filtering to remove clearly irrelevant works, such as papers that mentioned agents only tangentially or focused on unrelated domains like robotics or gaming. Next, we conducted abstract-based filtering to select papers that proposed, implemented, or benchmarked agentic systems supporting at least one stage of the data science lifecycle. To be included, a system had to \textbf{(i)} use an LLM as the primary reasoning component, \textbf{(ii)} support at least one stage of the data science lifecycle, such as business understanding, data acquisition, exploratory analysis, feature engineering, model development, interpretation or deployment, \textbf{(iii)} operate on structured data or code, and \textbf{(iv)} demonstrate either multi-step planning and tool orchestration or structured prompting that generated and executed analyses, queries or visualizations. We excluded papers that did not use LLMs (e.g., rule-based or classical machine learning systems), those restricted to domains like robotics or gaming, and those performing only static text generation without reasoning over data or code. Purely conceptual papers or interface prototypes without a functioning implementation were also excluded.

To ensure objectivity and transparency in the selection process, we adopted the PRISMA (Preferred Reporting Items for Systematic Reviews and Meta-Analyses) protocol~\cite{shamseer2015preferred}, which provides structured and reproducible guidelines for identifying, screening, and reporting relevant studies. This complements our inclusion and exclusion criteria and enhances the reproducibility of our methodology.

Title and abstract screening reduced the 587 papers to approximately 200 relevant papers. Full-text review of these 200 papers yielded 45 distinct DS agents. We analyzed these agents to extract their supported lifecycle stages, capabilities, challenges, design patterns, reasoning strategies, and evaluation methods. These findings informed the taxonomy introduced in Section~\ref{sec:Taxonomy}, which organizes the field along two complementary dimensions: coverage of the data science lifecycle and cross-cutting design attributes such as reasoning style, multimodality, and trust mechanisms.

Figure~\ref{fig:timeline_diag} provides a chronological overview of the emergence of agentic systems relevant to data science. Early foundational systems such as ReAct \cite{yao2023react} and AutoGPT \cite{yang2023autogpt} established core patterns for reasoning and tool use, which subsequently inspired specialized DS agents including InsightPilot \cite{sahu2024insightbench}, DatawiseAgent \cite{you2025datawiseagent}, and DS-Agent \cite{guo2024ds} that appeared between 2023 and 2025. This progression highlights the rapid acceleration of research in this area and motivates the lifecycle-based taxonomy.

% We conducted a systematic literature review to map and analyze research on agentic AI systems supporting data science workflows. Our search covered peer-reviewed articles, preprints, and conference proceedings published between 2023 and 2025, using sources such as Google Scholar and major AI venues (e.g., NeurIPS, ICLR, ACL, AAAI, TMLR). Search terms included \emph{data science agents}, \emph{LLM agents}, \emph{multimodal agents}, \emph{tool-using agents}, \emph{trustworthy AI systems}, and \emph{data science automation}. The review process involved two stages: an initial title-based screening to remove irrelevant works, followed by abstract-based filtering for finer selection. We included studies that proposed, implemented, or benchmarked agentic systems addressing any stage of the data science lifecycle. This process yielded approximately 200 papers, from which we identified and analyzed 45 distinct data science agents.

% \joty{This is very short to be a section by itself. We should merge it with Sec. 4.}

% \textcolor{red}{I will merge this with section 4}
% \enamul{I think it is still a good idea to have a separate method section but more elborated.Here is one example(Sec 2.1) although not perfect. https://onlinelibrary.wiley.com/doi/full/10.1111/cgf.15266}

% \joty{We should include groundbreaking work on agents like AutoGPT, ReAct in that timeline (Fig 2). I may have missed but make sure you include those that are pioneering in agents.}\saidul{updated}

\section{Taxonomy of Agentic AI Systems for Data Science}
\label{sec:Taxonomy}

% Recent advances have produced a growing variety of LLM-based agents for data science, yet their capabilities remain scattered across tasks, modalities, and design paradigms. Existing surveys typically emphasize individual systems or narrow use cases, making it difficult to compare agents, determine which stages they cover, or identify critical gaps. To address this, we derived a taxonomy by systematically reviewing representative agentic systems and identifying recurring patterns in both their functional roles and architectural designs. This dual structure captures the stages of the data science lifecycle that agents address and the ways in which they are designed to operate. Our taxonomy comprises two complementary components: (1) a lifecycle-driven classification aligned with key stages of the data science process, and (2) a set of cross-cutting design attributes that characterize how agents reason, interact, and align with human goals. Together, these components provide the first structured framework for mapping, comparing, and evaluating LLM-based data science agents in terms of both their functional scope and design choices.

LLM-based DS agents have emerged in diverse forms, yet their capabilities remain scattered across tasks, modalities, and design paradigms. Existing surveys typically emphasize individual systems or narrow use cases, making it difficult to compare agents, determine the stages they cover, or identify critical gaps. To address this, we propose a taxonomy with two components: (i) a lifecycle-driven classification aligned with the stages of the data science process, and (ii) cross-cutting design attributes that capture how agents reason, interact, and align with human goals. This framework provides a clear basis for mapping, comparing, and evaluating agent capabilities, limitations, and challenges.
\subsection{Lifecycle Stages (S1–S6)}
% \enamul{through out the paper I did not find any example of data science problem. Can you explain this workflow with a real world example?}
% Our taxonomy is organized around six core stages of the data science lifecycle \cite{cao2017data,boehm2019systemds}. These stages serve as the foundation for mapping and analyzing LLM-based agents in this survey:

Our taxonomy is organized around six core stages of the data science lifecycle \cite{cao2017data,boehm2019systemds}, derived by consolidating related activities from the broader end-to-end process, as illustrated in Figure~\ref{fig:ds-workflow}. These stages serve as the foundation for mapping and analyzing LLM-based agents in this survey:

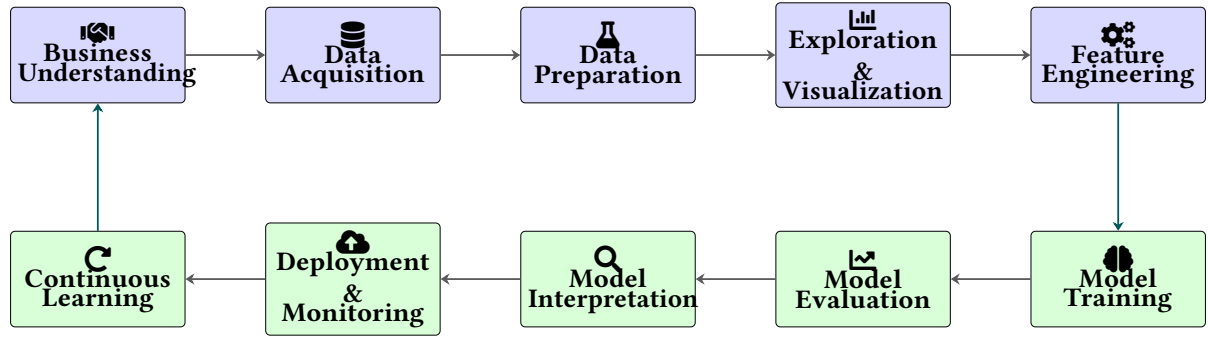
\begin{figure}[ht]
\centering
\resizebox{\linewidth}{!}{%
\begin{tikzpicture}[
  >={Stealth[length=4pt,width=5pt]},
  % Node styles with consistent height
  stageTop/.style = {
      draw, fill=blue!15, rounded corners=2pt,
      minimum height=1.5cm, minimum width=2.7cm,
      font=\footnotesize\bfseries, align=center,
      text width=2.5cm
  },
  stageBot/.style = {
      draw, fill=green!15, rounded corners=2pt,
      minimum height=1.5cm, minimum width=2.7cm,
      font=\footnotesize\bfseries, align=center,
      text width=2.5cm
  },
  % Node distance
  node distance = 2.2cm and 1.25cm
]
% -------- Top row --------
\node[stageTop] (bu) {\Large\faHandshake\\[-0.5em]Business\\[-0.1em]Understanding};
\node[stageTop, right=of bu] (da) {\Large\faDatabase\\[-0.5em]Data\\[-0.1em]Acquisition};
\node[stageTop, right=of da] (dp) {\Large\faFlask\\[-0.5em]Data\\[-0.1em]Preparation};
\node[stageTop, right=of dp] (ev) {\Large\faChartBar\\[-0.5em]Exploration \&\\[-0.1em]Visualization};
\node[stageTop, right=of ev] (fe) {\Large\faCogs\\[-0.5em]Feature\\[-0.1em]Engineering};
% -------- Bottom row -----
\node[stageBot, below=2.0cm of fe] (mt) {\Large\faBrain\\[-0.5em]Model\\[-0.1em]Training};
\node[stageBot, left=of mt]        (me) {\Large\faChartLine\\[-0.5em]Model\\[-0.1em]Evaluation};
\node[stageBot, left=of me]        (mi) {\Large\faSearch\\[-0.5em]Model\\[-0.1em]Interpretation};
% Use \faIcon with FA5 names for the two missing ones:
\node[stageBot, left=of mi]        (dm) {\Large\faIcon{cloud-upload-alt}\\[-0.5em]Deployment \&\\[-0.1em]Monitoring};
\node[stageBot, left=of dm]        (cl) {\Large\faIcon{redo-alt}\\[-0.5em]Continuous\\[-0.1em]Learning};

% Arrows
\foreach \a/\b in {bu/da, da/dp, dp/ev, ev/fe,
                   mt/me, me/mi, mi/dm, dm/cl}
  \draw[->, thick, gray!70!black] (\a) -- (\b);
\draw[->, thick, teal!70!black] (fe) -- (mt); % downward
\draw[->, thick, teal!70!black] (cl) -- (bu); % upward
\end{tikzpicture}%
}
\caption{End-to-End Data-Science Workflow.}
\label{fig:ds-workflow}
\end{figure}

\begin{itemize} [leftmargin=*, nosep] 
    \item \textbf{Business Understanding and Data Acquisition (S1)}: Agents in this category support problem formulation by translating business goals into analytical tasks, understanding domain requirements, and sourcing relevant data. Typical responsibilities include automated data extraction, cleaning, integration, quality checks, compliance validation, and early feature selection.

    \item \textbf{Exploratory Data Analysis and Visualization (S2)}: Agents in this stage assist with summarizing datasets, detecting anomalies, identifying patterns, and generating visualizations to help users understand data distributions and correlations.

    \item \textbf{Feature Engineering (S3)}: This stage includes agents that create new features, transform variables, and apply encoding or dimensionality reduction techniques to optimize data for modeling.

    \item \textbf{Model Building and Selection (S4)}: Agents in this phase automate model selection, training, and tuning, including hyperparameter optimization, model comparison, and performance evaluation. At this stage, fairness and bias mitigation are critical considerations, alongside techniques to prevent data leakage and ensure robust training.
    \item \textbf{Interpretation and Explanation (S5)}: Agents in this stage focus on explaining model predictions, generating actionable insights, and communicating results through interpretable outputs. This may involve techniques such as SHAP or LIME to highlight feature importance, provide counterfactual explanations, and produce narrative summaries that enhance transparency and user trust.
    \item \textbf{Deployment and Monitoring (S6)}: The final stage includes agents that deploy models into production, monitor performance, detect concept drift, fairness issues, or security risks, and trigger automated retraining or alerts. Compliance checks, audit logging, and governance controls are key for maintaining reliability and accountability in real-world environments.
\end{itemize}
Each agent is mapped to one or more lifecycle stages based on its primary functionality. This classification provides a structured view of how agentic systems support the data science process and reveals gaps, particularly in deployment, monitoring, and continuous feedback, where current maturity is limited.
\subsection{Cross-Cutting Design Attributes}
In addition to lifecycle classification, we identify five cross-cutting design attributes that offer a complementary perspective on how LLM-based agents reason, interact, orchestrate tools, learn, and ensure trust and safety. Unlike lifecycle stages, which capture what agents do, these attributes describe how they operate:

\begin{itemize} [leftmargin=*, nosep]
    \item \textbf{Reasoning and Planning Styles}: Agents differ in their reasoning approaches, ranging from simple single-step prompting to multi-step planning using either linear strategies (such as ReAct or chain-of-thought prompting) or hierarchical strategies (such as Tree-of-Thoughts or planner-coder-critic architectures) \cite{wang2024survey,yao2023react,wei2023chainofthoughtpromptingelicitsreasoning}. Some agents also incorporate self-reflective feedback loops to iteratively improve decision-making over time.

    \item \textbf{Modality Integration}: While some agents operate exclusively on text, others integrate multiple modalities such as text, tables, code, images, and visualizations. Multimodal capabilities are critical for complex data science tasks that involve heterogeneous data sources.

    \item \textbf{Tool Orchestration Depth}: Agents vary in the depth of their external tool use, ranging from no tool use to light orchestration (such as calculators or simple APIs) to deep orchestration involving code execution, database querying, multi-agent collaboration, or the use of orchestration frameworks such as LangChain, LlamaIndex, or AutoGen to manage complex toolchains and interactions \cite{shen2023hugginggpt,yao2023react,ugwueze2024continuous,langgraph2024}.  These frameworks provide abstractions for tool invocation, memory management, inter-agent messaging, and multi-step task planning, making them common building blocks in modern DS agents.
    
    \item \textbf{Learning and Alignment Paradigms}: Agents employ diverse strategies, including zero-shot prompting, few-shot learning, and instruction tuning \cite{ouyang2022training,hassan2023chatgpt}. More advanced approaches incorporate RL from human feedback (RLHF) \cite{christiano2017deep}, direct preference optimization (DPO) \cite{rafailov2023direct}, and emerging methods such as Group Relative Policy Optimization (GRPO) \cite{wei2025swerl,liu2024deepseek}.

    \item \textbf{Trustworthiness and Safety Mechanisms}: Agents may include safeguards for fairness, explainability, privacy, robustness, and human oversight \cite{zhao2024explainability}. They also implement risk-mitigation mechanisms against hallucination and unsafe tool use, which are critical for reliable deployment in real-world applications \cite{shen2023hugginggpt,yehudai2025survey}.  
\end{itemize}

To ground these stages in a real-world context, consider an online banking fraud detection project. In \textbf{S1}, the business goal is to reduce missed fraud while avoiding excessive false positives, which requires identifying suspicious patterns and building a classifier. This also involves securely acquiring relevant transactional and customer data. In \textbf{S2}, the raw data are cleaned and explored to reveal transaction patterns. In \textbf{S3}, domain informed features such as average transaction amount, login frequency, and geographic movement are engineered to highlight anomalies. In \textbf{S4}, models are trained and evaluated with metrics like precision and recall to capture the cost of fraud errors. In \textbf{S5}, interpretability tools explain why a transaction was flagged, aiding compliance and customer communication, while dashboards summarize key findings and support daily fraud analysis. Finally, in \textbf{S6}, the code is packaged for deployment with security measures to prevent adversarial attacks, and the deployed model is monitored for drift and continuously updated to counter evolving attack patterns. Throughout, the cross-cutting attributes of \emph{trustworthiness}, \emph{alignment}, and \emph{safety} ensure the agentic system operates effectively, ethically, and securely in this financial domain.

\
\section{Agentic Capabilities Across the Data Science Lifecycle}

\label{sec:lifecycle} 
Building on the taxonomy introduced in Section~\ref{sec:Taxonomy}, this section analyzes how the 45 identified DS agents align with the six stages of the data science lifecycle. For each stage, we examine representative systems, their core tasks, and available benchmarks, highlighting stage-specific strengths, gaps, and challenges to provide a structured view of current capabilities and opportunities for advancing end-to-end automation. Table~\ref{tab:agents} presents a consolidated summary of these 45 agents, annotated across lifecycle stages, reasoning and planning strategies, modalities, tool use, learning and alignment methods, and trust and safety mechanisms. This overview serves as a reference point for the stage-wise analysis that follows.

% Building on the taxonomy introduced in Section~\ref{sec:Taxonomy}, this section analyzes how the 45 data science agents identified in our review support the six core stages of the data science lifecycle. For each stage, we examine which agents contribute to that stage, the types of tasks they perform, their key functionalities, and representative frameworks, referencing benchmarks where available. We also map the breadth of coverage across stages to highlight which phases are well supported and which remain underrepresented. Finally, we summarize major trends, common limitations, and open challenges, providing a stage-by-stage view of current capabilities, remaining gaps, and opportunities for advancing data science agents toward end-to-end automation.

\definecolor{headerblue}{RGB}{226,239,255}
\definecolor{rowgray}{gray}{0.96}

\begin{table}[htbp]
\scriptsize
\centering
\caption{Taxonomy of Data Science Agents: Lifecycle Stages, Capabilities,  and Design Dimensions. \footnotesize Stage: S\# = lifecycle, NL = non-lifecycle}
\setlength{\tabcolsep}{3pt}
\rowcolors{2}{rowgray}{white} % alternate row shading
\begin{tabular}{|p{2.2cm}|p{0.6cm}|p{1.3cm}|p{2.1cm}|p{3.0cm}|p{1.9cm}|p{2.2cm}|}

% \begin{tabular}{|p{2.5cm}|p{1.1cm}|p{2.1cm}|p{1.8cm}|p{2.3cm}|p{2.1cm}|p{2.1cm}|}
\hline
\rowcolor{headerblue}
\textbf{Agent Name} & \textbf{Stage} & \textbf{Reasoning \& Planning} & \textbf{Modality} & \textbf{Tool Use} & \textbf{Learning \& Alignment} & \textbf{Trust \& Safety} \\
\hline

SCALE \cite{zhaoscale} % [June 2025]
& S2 & Hierarchical & Text & None & Human-in-loop & Bias checks, manual review \\

SPIO \cite{seo2025spio} % [30 Mar 2025]
& S2-S4 & Linear & Tabular & Python, APIs & Prompt-only & None \\

DatawiseAgent \cite{you2025datawiseagent} % [10 Mar 2025]
& S2-S4 & Hierarchical & Text, Tabular & Python, GPT-4o-mini visual & Prompt-only, feedback-driven & None \\

AgentSafe \cite{mao2025agentsafe} % [6 Mar 2025]
& NL & Hierarchical & Text, Code, Tabular & LLM API calls & Prompt-only & Built-in guard rails \\

OpenCHA-PPG \cite{feli2025llm} % [18 Feb 2025]
& S2-S4 & Hierarchical & Text, Tabular & External signal processing tools & Prompt-only & Transparent explanation \\

Harmonia \cite{santos2025interactive} % [10 Feb 2025]
& S1 & Hierarchical & Text, Tabular & bdi-kit, Python, schema & Prompt-only & Partially addressed \\

PlotGen \cite{goswami2025plotgen} % [3 Feb 2025]
& S2 & Hierarchical & Text, Code, Tabular & Python, GPT-4V & Prompt-only & None \\

Jupybara \cite{wang2025jupybara} % [28 Jan 2025]
& S2, S5 & Linear & Text, Code, Tabular & Jupyter Notebooks, Python & Prompt-only & None \\

XMODE \cite{nooralahzadeh2024explainable} % [24 Dec 2024]
& S2-S5 & Linear & Text, Tabular, Images & SQL queries, image analysis modules, Python & Prompt-only & Explainability, transparent re-planning \\

Star-Agents \cite{zhou2024star} % [21 Nov 2024]
& S4 & Hierarchical & Text & LLM API calls & Prompt-only & None \\

Agent K v1.0 \cite{grosnit2024large} % [5 Nov 2024]
& S2-S4 & Linear & Text, Tabular, Image & AutoML frameworks, Bayesian optimization, Kaggle APIs & Experience based learning & None (Unit Test) \\

AutoKaggle \cite{li2024autokaggle} % [27 Oct 2024]
& S1-S4 & Linear & Text, Tabular & ML Tools Library, Kaggle API & Prompt-only, unit tests & Unit tests, plan reviews \\

LLM-MAS \cite{de2024can} % [25 Oct 2024]
& NL & Linear & Text, Code & LLM API calls only & Prompt-only & Four trustworthiness techniques \\

GeoAgent \cite{chen2024llm} % [24 Oct 2024]
& S1-S2 & Linear & Text, Code, Geodata & Python, Geospatial APIs & Prompt-only & Human-in-loop \\

DocETL \cite{shankar2024docetl} % [16 Oct 2024]
& S2-S3 & Hierarchical & Text & Python, Rust engines & Prompt-only & LLM validators \\

AutoML-Agent \cite{trirat2024automl} % [3 Oct 2024]
& S1-S4& Hierarchical & Text, Tabular, Images & Retrieval augmented, Python & Prompt-only & Verification only \\

Data Formulator 2 \cite{wang2025data} % [28 Aug 2024]
& S2, S5 & Basic I/O & Text, Tabular, Code & Vega-Lite & Prompt-only & Explanations \\

WaitGPT \cite{xie2024waitgpt} % [3 Aug 2024]
& S2 & Linear & Text, Code, Visuals & Python, AST & Prompt-only & User verifiable \\

LAMBDA \cite{maojun2025lambda} % [24 Jul 2024]
& S2-S4 & Basic I/O & Text, Tabular, Code & Python, ML Modeling & Prompt-only & Human-in-loop \\

CellAgent \cite{xiao2024cellagent} % [13 Jul 2024]
& S2-S4 & Hierarchical & Text, Tabular & Biological tools & Prompt-only & None \\

AgentPoirot \cite{sahu2024insightbench} % [8 Jul 2024]
& S1-S5 & Hierarchical & Text, Tabular, Code & Python & Prompt-only & None \\

TABLELLM \cite{zhang2024tablellm} % [28 Mar 2024]
& S2 & Linear & Text, Tabular, Visual & Python & Fine-tuned & Cross-way validation \\

CleanAgent \cite{qi2024cleanagent} % [13 Mar 2024]
& S2 & Linear & Text, Tabular & Dataprep Clean’s Functions & Prompt-only & Automated verification \\

WebVoyager \cite{he2024webvoyager} % [25 Jan 2024]
& NL & Hierarchical & Text, Visual, Code & Python & Prompt-only & Browser environment \\

Data Interpreter \cite{hong2024data} % [28 Feb 2024]
& S2-S5 & Hierarchical & Text, Tabular, Code & Python & Prompt-only & None \\

DS-Agent \cite{guo2024ds} % [27 Feb 2024]
& S2-S6 & Linear & Text, Tabular, Code & Python, API calls & Prompt-only & None \\

MatPlotAgent \cite{yang2024matplotagent} % [18 Feb 2024]
& S2 & Linear & Text, Code, Visuals & Python & Prompt-only & None \\

LLMDB \cite{zhou2024llm} % [4 Feb 2024]
& S2 & Hierarchical & Text, Tabular & Vector databases, external APIs & Prompt-only & None \\

QDA-MultiAgent \cite{rasheed2024can} % [2 Feb 2024]
& S2 & Hierarchical & Text, Documents & Python & Prompt-only & None \\

TROVE \cite{wang2024trove} % [23 Jan 2024]
& S2 & Hierarchical & Text, Tabular & Python, SQL & Prompt-only & None \\

DAAgent \cite{hu2024infiagent} % [10 Jan 2024]
& S2-S4 & Hierarchical & Text, Tabular, Code & Python & Instruction tuned & None \\

TAP4LLM \cite{sui2023tap4llm} % [14 Dec 2023]
& S2 & Hierarchical & Text, Tabular & SQL & Prompt-only & None \\

Toolformer \cite{schick2023toolformer} % [10 Dec 2023]
& S2 & Hierarchical & Text, Tabular & API calls, Calculator, Calendar & Self-supervised & None \\

JarviX \cite{liu2023jarvix} % [3 Dec 2023]
& S2-S4 & Hierarchical & Text, Tabular, Audio & Postgres, ElasticSearch, FAISS, AutoML & Prompt-only & Human-in-loop \\

LLM4Vis \cite{wang2023llm4vis} % [11 Oct 2023]
& S2 & Hierarchical & Text & API calls & Prompt-only & None \\

Data Formulator \cite{wang2023data} % [18 Sep 2023]
& S2 & Hierarchical & Text, Tabular & LLM's API & Prompt-only & Validation guard rails \\

AutoGen \cite{wu2023autogen} % [16 Aug 2023]
& S2-S3 & Hierarchical & Text, Code, Tabular & External tool integration, APIs, databases & Prompt-only & Code-safety checks, Human-in-loop \\

Data-Copilot \cite{zhang2023data} % [12 Jun 2023]
& S2 & Hierarchical & Text, Tabular & Python, Matlab & Prompt-only & None \\

GPT-4 Analyst \cite{cheng2023gpt} % [24 May 2023]
& S2-S3 & Basic I/O & Text, Tabular & Python, SQL & Prompt-only & None \\

VIDS \cite{hassan2023chatgpt} % [23 May 2023]
& S1-S4 & Linear & Text & ChatGPT, Scikit Learn & Prompt-only & None \\

CAAFE \cite{hollmann2023large} % [5 May 2023]
& S3-S4 & Hierarchical & Text, Tabular, Code & LLMs calls, ML Models & Prompt-only & Human-in-loop \\

InsightPilot \cite{ma2023demonstration} % [2 Apr 2023]
& S2 & Hierarchical & Text, Tabular, Code & Three insight engines & Prompt-only & None \\

HuggingGPT \cite{shen2023hugginggpt} % [30 Mar 2023]
& NL & Hierarchical & Text, Vision, Speech, Video & ChatGPT, Hugging Face models & Prompt-only & None \\

Chat2VIS \cite{maddigan2023chat2vis} % [24 Mar 2023]
& S2 & Linear & Text & LLMs call, Python & Prompt-only & None \\

LIDA \cite{dibia2023lida} % [6 Mar 2023]
& S2 & Linear & Text & LLMs call, Diffusion models & Prompt-only & Self evaluation \\

\hline
\end{tabular}
\label{tab:agents}
\end{table}

\subsection{Business Understanding and Data Acquisition}

Translating high-level business goals into concrete analytical questions remains one of the most persistent bottlenecks in enterprise data workflows. At this early stage, DS agents are expected to bridge the gap between ambiguous objectives and actionable tasks, yet this capability is still underdeveloped. Effective progress in this stage is critical, since errors in problem formulation or data acquisition propagate downstream, reducing reliability across the entire pipeline. 

Several systems and benchmarks have begun addressing this challenge. InsightBench~\cite{sahu2024insightbench} is among the first benchmarks to evaluate an agent’s ability to translate business goals into analytical tasks, requiring multi-step reasoning that spans question formulation, interpretation, and actionable summarization. Its baseline system, AgentPoirot (GPT-4o), recovers about 60\% of predefined insights under well-specified goals but only around 40\% when objectives are vague or open-ended, highlighting the difficulty of reasoning from under-specified prompts. Similarly, AutoKaggle~\cite{li2024autokaggle} incorporates business understanding into automated workflows, linking initial goal formulation with early steps such as data cleaning and exploratory analysis. On the data acquisition side, recent LLM-based agents~\cite{piccialli2025agentai,du2024contextaware,zhang2024benchmarking} have improved schema parsing, anomaly detection, and cleaning for structured and semi-structured formats. These systems can also extract relevant information from unstructured sources such as contracts, support tickets, and policy documents, transforming them into structured inputs for analysis and model development~\cite{piccialli2025agentai,du2024contextaware,zhang2024benchmarking}. 

Despite this progress, real-world enterprise environments remain far more complex. They involve nested schemas, multi-table relationships, and heterogeneous workflows that often combine dashboards with command-line interfaces. Benchmarks highlight these difficulties. On Spider 1.0~\cite{yu2018spider}, GPT-4 based agents achieve around 86\% accuracy on single-query text-to-SQL tasks, but performance drops sharply on multi-relational queries in Spider 2.0~\cite{lei2024spider}, where even GPT-4o achieves only about 10\% accuracy~\cite{gao2023text,deng2025reforce}. Spider2-V~\cite{cao2024spider2}, which tests end-to-end execution across mixed interfaces, reports success rates below 14\%, highlighting orchestration challenges in ETL pipelines that integrate tools like Airbyte and BigQuery. ELT-Bench~\cite{jin2025elt} further stresses multi-step ingestion workflows, requiring connector configuration, heterogeneous format transformation, and SQL model generation across platforms such as Snowflake. Across these evaluations, current DS agents fall short of enterprise-grade reliability in goal translation and data integration.

Beyond technical accuracy, this stage raises crucial issues around privacy, fairness, and compliance. Improper handling of sensitive attributes or confidential data can propagate bias, introduce security risks, or create legal liabilities. Mature enterprise pipelines mitigate these risks through safeguards such as schema validation, drift detection, feature-importance audits, leave-one-feature-out retraining, anonymization or pseudonymization, and synthetic data generation for testing. For DS agents to reliably support business understanding and data acquisition, these safeguards must be embedded into agentic workflows from the outset. Future systems should therefore combine adaptive reasoning, transparent decision-making, and integrated compliance checks to align with both organizational goals and regulatory standards~\cite{zeng2025alignment}. As a result, limitations at this stage remain a key barrier to achieving robust, trustworthy, and end-to-end automation in the data science lifecycle.

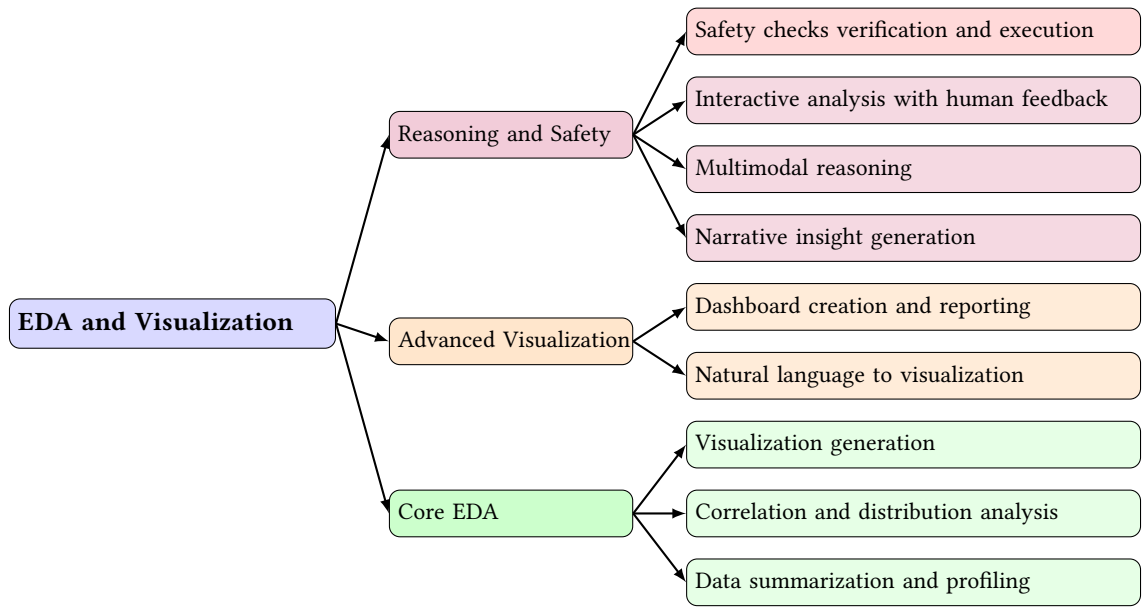
\begin{figure}[ht]
\centering
\begin{forest}
  for tree={
    draw,
    rounded corners,
    align=center,
    font=\small,
    edge={-latex, thick},
    grow=0,
    parent anchor=east,
    child anchor=west,
    l sep=20pt,
    s sep=8pt,
    text width=5.8cm, % default for children
    minimum height=0.4cm
  }
  [EDA and Visualization, fill=blue!15, text width=4.1cm, font=\normalsize\bfseries
    [Core EDA, fill=green!20, text width=3.0cm % reduced width for category
      [Data summarization and profiling, fill=green!10]
      [Correlation and distribution analysis, fill=green!10]
      [Visualization generation, fill=green!10]
    ]
    [Advanced Visualization, fill=orange!20, text width=3.0cm % reduced width for category
      [Natural language to visualization, fill=orange!15]
      [Dashboard creation and reporting, fill=orange!15]
    ]
    [Reasoning and Safety, fill=purple!20, text width=3.0cm % reduced width for category
      [Narrative insight generation, fill=purple!15]
      [Multimodal reasoning, fill=purple!15]
      [Interactive analysis with human feedback, fill=purple!15]
      [Safety checks verification and execution, fill=red!15]
    ]
  ]
\end{forest}
\caption{EDA and Visualization capabilities of DS agents, organized into three layers: core EDA, advanced visualization, and reasoning and safety aspects.}

\label{fig:eda-viz-capabilities}
\end{figure}

\subsection{Exploratory Data Analysis and Visualization}

Exploratory Data Analysis (EDA) is a critical stage for identifying trends, anomalies, and relationships that guide hypothesis generation and model design. Recent advances in LLM-powered agents have enabled partial automation of EDA by integrating natural language understanding with code execution and visualization capabilities \cite{zhang2024benchmarking}. These systems can translate natural queries into executable analyses and visual reports, offering interactive support for data exploration. However, their effectiveness remains limited, with challenges in structured planning, domain-aware reasoning, multimodal synthesis, and trustworthy reporting. Figure~\ref{fig:eda-viz-capabilities} summarizes the key capability layers we identify in this stage: (i) core EDA tasks, (ii) advanced visualization, and (iii) reasoning, which encompasses multimodal integration, interactive analysis, and safety mechanisms. We use this structure to organize our discussion of representative systems below.

% Exploratory Data Analysis (EDA) is a critical stage for identifying trends, anomalies, and relationships that guide hypothesis generation and model design.  Recent advances in LLM-powered agents have enabled partial automation of EDA by integrating natural language understanding with code execution and visualization capabilities \cite{zhang2024benchmarking}. These systems can translate user queries into executable Python or SQL code, compute descriptive statistics, generate plots, and produce narrative interpretations, often within interactive notebook environments. 

% Benchmarks such as InsightBench highlight both progress and persistent limitations: while AgentPoirot (GPT-4o) recovers about 60\% of predefined insights under well-specified goals, performance drops to roughly 40\% for vague or open-ended tasks \cite{sahu2024insightbench}. These results highlight the need for structured planning, contextual interpretation, and domain-aware reasoning, alongside advances in multimodal synthesis, interactive exploration, and trustworthy reporting. 

\subsubsection{\textbf{Domain Knowledge and Context-Aware Reasoning}}

Domain-aware reasoning is critical in EDA because agents must interpret datasets within context, apply appropriate methods, and avoid outputs that are statistically correct but analytically irrelevant. For example, an agent might accurately compute correlation coefficients yet fail to recognize that variables represent different measurement units, yielding spurious associations.  Benchmarks reveal significant limitations in domain reasoning. BrowseComp~\cite{wei2025browsecomp} evaluates factual retrieval and multi-step reasoning over domain-specific datasets. Even browser-enabled GPT-4o agents complete fewer than 2\% of tasks requiring external information retrieval, navigation, and synthesis. While retrieval-augmented generation (RAG) offers one pathway to improve grounding by conditioning outputs on external knowledge bases~\cite{zhao2024retrieval}, current implementations remain brittle. Schema mismatches, unit inconsistencies, and noisy sources frequently produce outputs that appear coherent but lack analytical validity~\cite{piccialli2025agentai}.  

Domain-specific interpretation failures are particularly evident in specialized fields. In finance, agents misinterpret temporal dependencies in time-series data~\cite{piccialli2025agentai}, while in healthcare and climate science, they apply inappropriate statistical tests to non-normal distributions~\cite{zhang2024benchmarking}. These errors propagate downstream: flawed EDA assumptions yield invalid features and biased models. Future progress requires domain-adaptive benchmarks, retrieval systems robust to schema drift, and validation checks that flag questionable inferences. Without reliable domain grounding at this stage, downstream modeling and decision-making remain fundamentally constrained.

\subsubsection{\textbf{Data Visualization}} 
Visualization transforms analytical results into interpretable formats that support exploration and decision-making. Without reliable visualization, even correctly computed statistics may fail to guide meaningful insights.

Early systems demonstrated that LLMs can lower barriers to chart creation by generating code directly from natural language. Chat2VIS~\cite{maddigan2023chat2vis} outperformed rule-based interfaces by flexibly translating queries into  executable code. LIDA~\cite{dibia2023lida} and Prompt4Vis~\cite{li2024prompt4vis} extended this with schema-aware and grammar-agnostic prompting, while ChartGPT~\cite{tian2024chartgpt} improved alignment between abstract analytical goals and visual design. More recent frameworks emphasize iterative refinement and multi-agent collaboration. METAL~\cite{li2025metalmultiagentframeworkchart} assigns specialized agents to code review and visual verification, surpassing GPT-4 baselines by over 11\% on chart-to-code benchmarks. PlotGen~\cite{goswami2025plotgen} incorporates multimodal feedback for scientific visualization,  adjusting visual encodings based on rendered outputs. DatawiseAgent~\cite{you2025datawiseagent} integrates planning and self-debugging in notebook environments, and Text2Vis~\cite{text2vis2025} applies a cross-modal actor–critic framework that improves GPT-4o’s pass rate from 26\% to 42\%. Interactive approaches such as WaitGPT~\cite{xie2024waitgpt} enable real-time human guidance, while LLM4Vis~\cite{wang2023llm4vis} pairs charts with natural language rationales to strengthen transparency. Beyond single-chart generation, recent systems such as Data-to-Dashboard~\cite{zhang2025datatodashboardmultiagentllmframework} and EduVisAgent~\cite{ji2025eduvisbencheduvisagentbenchmarkmultiagent} extend visualization to multi-view dashboards, coordinating domain detection, concept extraction, and visualization planning. Evaluations on EduVisBench confirm that these multi-agent pipelines deliver deeper insights and stronger domain relevance than single-agent baselines.

Despite this progress, visualization agents exhibit three persistent failure modes \cite{text2vis2025}. First, semantic grounding errors: agents select inappropriate chart types such as pie charts for temporal trends or apply logarithmic scales to count data, distorting interpretation. Second, visual fidelity issues: outputs contain truncated labels, overlapping annotations, or missing legends that render charts unreadable without manual correction. Third, intent misalignment: generated visualizations may be syntactically correct but fail to address the user's analytical goal, such as producing distribution plots when comparative analysis is needed~\cite{text2vis2025}. Multi-agent architectures like METAL mitigate but do not eliminate these issues. While verification agents catch some errors, they introduce latency and cost overheads. More fundamentally, benchmark-driven improvements risk overfitting to common patterns while failing on novel domain-specific visualizations. Addressing these limitations requires deeper integration of data semantics and visual design principles, alignment techniques that model user intent rather than optimize generic correctness, and evaluation frameworks that assess interpretability and decision utility in realistic workflows. DS agents must progress from producing syntactically valid charts to generating visualizations that are faithful to data, aligned with user goals, and robust across domains.

\subsubsection{\textbf{Multimodal Reasoning and Fusion in EDA}}
\label{sec:multimodal}

Exploratory data analysis rarely involves text alone: analysts work with tables, charts, documents, code, and dashboards in parallel. Text-only agents often misinterpret variable semantics or fail to link visual features with underlying data, producing outputs that look plausible but lack analytic validity~\cite{chartqapro, xie2024osworld, wei2025swe, bigdocs, tablevqa, longfin}.  Multimodal reasoning addresses these limitations by enabling DS agents to integrate heterogeneous sources and produce outputs that are both technically sound and contextually relevant. 

Work in this area spans three dimensions. First, input-focused benchmarks test whether agents can parse diverse DS artifacts: ChartQA~\cite{chartqapro} targets chart reasoning, DocVQA~\cite{docvqa} scanned documents, TableVQA~\cite{tablevqa} tabular data, OSWorld~\cite{xie2024osworld} and ScreenSpot~\cite{screenspot} UI-based dashboards, and SWE-Bench~\cite{wei2025swe} code reasoning in GitHub issues. These benchmarks have driven progress but remain narrow, emphasizing isolated tasks rather than end-to-end workflows.
Second, output-oriented systems extend beyond text answers. Text2Vis~\cite{text2vis2025} translates natural language into executable visualization code, while multi-agent pipelines such as Data-to-Dashboard~\cite{zhang2025datatodashboardmultiagentllmframework} and EduVisAgent~\cite{ji2025eduvisbencheduvisagentbenchmarkmultiagent} automate dashboard generation with explanatory narratives, showing higher domain relevance on EduVisBench. Some systems explore richer outputs like images~\cite{gpt4-I} or videos~\cite{sora, opensora, veo3}, though their value for Data Science lies mainly in reporting and communication rather than core analysis.
Third, fusion architectures determine how modalities are integrated. Early fusion (e.g., LLaVA~\cite{llava}, Paligemma~\cite{paligemma}, AlignVLM~\cite{alignvlm}, Ovis~\cite{ovis}, LLaVA-Next~\cite{llavanext}) projects inputs into a shared representation, enabling tight cross-modal grounding but risking noise propagation. Late fusion (e.g., CLIP~\cite{clip}, SigLIP~\cite{siglip}, MIEB~\cite{xiao2025miebmassiveimageembedding}) encodes modalities independently, offering robustness for retrieval tasks but weaker joint reasoning. Cross-modal attention (e.g., Flamingo~\cite{flamingo}, LLaMA-3-Vision~\cite{llama3}) allows dynamic interaction, though at significant computational cost. On the reasoning side, Chain-of-Thought~\cite{wei2023chainofthoughtpromptingelicitsreasoning} and Program-of-Thought~\cite{chen2023programthoughtspromptingdisentangling} improve step-wise reasoning, while RL-trained methods such as O1~\cite{openai2024openaio1card} and DeepSeek-R1~\cite{deepseekai2025deepseekr1incentivizingreasoningcapability} refine reasoning traces. The emerging Visualization-of-Thought paradigm~\cite{li2025imaginereasoningspacemultimodal} is particularly promising for DS, embedding reasoning in vision space by generating intermediate visualizations.

Despite clear progress, multimodal DS agents remain brittle. Alignment errors such as mismatches between chart labels and table headers undermine grounding~\cite{alignvlm}. Token length explosion from large tables or high-resolution dashboards inflates costs; mitigation strategies such as pooling~\cite{hu2024mplugdocowl15unifiedstructure, swin}, fixed query tokens~\cite{flamingo}, or pixel shuffling~\cite{smolvlm} reduce compute but often discard important detail. Perception errors persist, with agents misreading scales, legends, or annotations~\cite{chartgemma, chartinstruct}, and OCR inaccuracies degrade document reasoning~\cite{docvqa}, pointing to the need for stronger vision encoders~\cite{clip, siglip, dinov2, alignvlm}. Visual reasoning itself remains underdeveloped: frameworks like Visual Sketchpad~\cite{hu2024visualsketchpadsketchingvisual}, ReFocus~\cite{fu2025refocusvisualeditingchain}, and perception tokens~\cite{bigverdi2024perceptiontokensenhancevisual} show promise, but integration into interactive EDA pipelines is limited. Finally, benchmark overfitting is a critical barrier: systems such as METAL~\cite{li2025metalmultiagentframeworkchart} report strong chart-to-code gains but fail to generalize to heterogeneous enterprise dashboards, where retrieval, multimodal alignment, and contextual reasoning are essential~\cite{chartqapro}.

In sum, multimodal reasoning has expanded the scope of what DS agents can automate, but real-world reliability is constrained by weak alignment, scalability issues, and narrow evaluation settings. Closing these gaps will require robust retrieval-enhanced fusion, benchmarks that reflect enterprise complexity, and safety checks that guarantee faithfulness across modalities. Without such safeguards, multimodal outputs risk being visually polished yet analytically misleading, eroding trust in EDA pipelines.

\subsubsection{\textbf{Interactive Analysis and Human–Agent Collaboration}} 

Interactive analysis enables agents and human analysts to collaboratively refine EDA outputs, clarify objectives, and adapt analyses to evolving needs. Rather than pursuing full automation, these systems combine natural language interfaces with direct manipulation: users specify high-level goals, while agents execute underlying data transformations.

Data Formulator~\cite{wang2023data} introduced concept binding, where users describe relationships in natural language and the system generates corresponding charts. Data Formulator 2~\cite{wang2025data} extended this with mixed-initiative interaction, iterative refinement, and history navigation for more efficient exploration. Other systems emphasize tighter feedback loops: WaitGPT~\cite{xie2024waitgpt} allows real-time monitoring and intervention, while DataPilot~\cite{narechania2023datapilot} supports visual data preparation and subset selection through quality and usage aware interaction design. These mixed-initiative pipelines leverage human domain expertise alongside automated execution.

Interactive EDA systems struggle with maintaining coherence across multi-turn interactions, preserving context during task switches, and avoiding the reintroduction of outdated operations \cite{wang2025data}. Addressing these gaps requires long-context state tracking and versioned revision histories that make analytic changes transparent. Stronger memory mechanisms and alignment with analyst intent will be critical for making collaboration both reliable and decision-ready.

% Interactive analysis enables agents and human analysts to collaboratively refine EDA outputs, clarify objectives, and adapt analyses to evolving needs. Rather than fully automating chart creation, these systems combine natural language interfaces with direct manipulation, allowing users to set high-level goals while agents handle data transformations.  Data Formulator \cite{wang2023data} introduces a “concept binding” approach where users describe relationships in natural language, and the system automatically generates corresponding charts. Data Formulator 2  \cite{wang2025data} extends this with mixed-initiative interaction, iterative refinement, and history navigation for efficient exploration. Other frameworks such as WaitGPT \cite{xie2024waitgpt} enable real-time monitoring, while DataPilot \cite{narechania2023datapilot} supports visual data preparation and subset selection through quality and usage-aware interaction design. 

% %\paragraph{\textbf{Key Limitations.}}

% \textcolor{black}{Despite progress, interactive EDA remains constrained by fragile narrative coherence across mixed-initiative turns, brittle context persistence under task switching, and the recurrent reintroduction of deprecated operations. Mitigating these limitations requires robust long-context state models that track analytic commitments and expose transparent, versioned revision rationales.}

\subsubsection{\textbf{Reporting and Insight Generation}}

Narrative generation complements visualization by translating analytical results into interpretable, actionable insights. Multi-agent frameworks increasingly automate this process through structured reasoning pipelines. The hypothesis-driven approach proposed in~\cite{perez2025llm} combines a Hypothesis Generator, Query Agent, and Summarization module to iteratively generate domain-relevant questions, execute SQL queries, and verbalize insights. On public and enterprise databases, this approach achieves significantly higher insightfulness scores while maintaining correctness compared to GPT-4 baselines. Visualization systems have begun integrating narrative reporting as a core feature. Data-to-Dashboard~\cite{zhang2025datatodashboardmultiagentllmframework} pairs interactive dashboards with structured commentary, emulating an analyst's interpretation through captions, summaries, and reports. By bridging technical outputs with accessible narratives, such systems help decision-makers without deep statistical expertise interpret results more effectively.

Despite this progress, reporting pipelines remain vulnerable to hallucinated claims, weak schema or ontology alignment, and templated prose that hides uncertainty and provenance. Addressing these issues requires evidence-linked grounding with explicit claim–evidence mappings, uncertainty quantification calibrated to analytic confidence, and human validation mechanisms aligned with domain-specific governance and risk tolerances. With these safeguards in place, narrative reporting can deliver trustworthy, decision-ready insights.

\subsubsection{\textbf{Trust and Safety in EDA}} 
Factual accuracy is essential in automated EDA, as hallucinated insights can mislead decisions and erode trust. Detection methods validate outputs against reference data, uncertainty measures, or retrieved evidence. Retrieval-based verification such as HalluMeasure~\cite{akbar2024hallumeasure} classifies statements as supported, contradicted, or absent. Ask-EDA~\cite{shi2024ask} achieves significant recall improvements on three domain-specific datasets (design QA, command handling, and abbreviation resolution) by combining hybrid RAG with abbreviation correction. Additional techniques include entropy-based uncertainty estimation~\cite{farquhar2024detecting}, calibrated trust scores via TLM~\cite{athalye2024overcoming}, ensemble-based factuality evaluation with FEWL~\cite{wei2024measuring}, and adversarial testing through benchmarks like Phare~\cite{dora2025hallucination}.

Mitigation strategies emphasize grounding, verifiability, and resilience to unsupported outputs. RAG conditions responses on authoritative sources, reducing hallucinations by anchoring generations in external knowledge. Ask-EDA~\cite{shi2024ask} enhances domain-specific accuracy through sparse–dense retrieval combined with abbreviation correction. Execution-based frameworks such as LightVA~\cite{zhao2024lightva} enforce correctness by generating and running Python or SQL code before producing final results, often paired with self-correction loops and chain-of-thought reasoning. Multi-agent systems like Jupybara~\cite{wang2025jupybara} further improve reliability by assigning separate planning and review agents to validate results iteratively. 

However, current implementations remain incomplete. Hallucination detection and mitigation techniques are rarely integrated into production EDA systems, and existing benchmarks test isolated components rather than end-to-end reliability. Closing this gap requires safety-aware system design and evaluation frameworks that reflect real-world analytical complexity.

\subsection{Feature Engineering (FE)}

Feature engineering (FE) transforms raw variables into informative representations that improve predictive performance. LLM-enabled agents extend this process by operationalizing EDA-derived hypotheses into concrete transformations, feature constructions, and selections. Without effective feature engineering, downstream models risk overfitting, redundancy, or reliance on spurious correlations.

Current research spans three directions. First, domain-driven feature construction integrates expert knowledge beyond standard transformations. LLM-FE~\cite{abhyankar2025llm} identified cholesterol as predictive in heart disease and constructed log-transformed features with medical justification. CAAFE~\cite{hollmann2023large} emphasized interpretability by pairing generated features with natural language explanations, though this creates tradeoffs between domain accuracy and user transparency. Multi-agent systems like AutoKaggle~\cite{li2024autokaggle} integrate these capabilities into end-to-end competition workflows. Second, automated feature selection combines statistical filtering with model-based scoring. AltFS~\cite{jia2024altfs} uses LLM-generated semantic rankings refined by deep recommenders to address redundancy and sparsity. Dynamic pipelines~\cite{zhang2024dynamic} move beyond one-shot engineering through iterative LLM prompts that refine features over time, improving cross-domain applicability. Third, RL approaches frame feature engineering as sequential decision-making. SMART~\cite{bouadi2024semantic} leverages knowledge graphs to generate interpretable features while balancing accuracy and complexity. Prompt-chaining methods enable agents to propose transformations, evaluate their impact, and update feature sets dynamically~\cite{khurana2018feature}.

Despite this progress, FE agents remain fragile. Generated features are often semantically plausible but statistically weak, contributing noise rather than signal. For example, an agent might create interaction terms between unrelated variables that increase training accuracy but fail on held-out data. Domain transfer is limited: LLM-FE performs well in healthcare but struggles in finance or climate domains~\cite{abhyankar2025llm}. Unlike EDA, which benefits from benchmarks such as InsightBench and MatplotBench, feature engineering lacks systematic evaluation beyond downstream task accuracy. Agents also increasingly depend on external tools and RAG to enrich feature spaces, but these integrations are brittle and amplify schema mismatches or noisy inputs~\cite{farooq2024survey}. Safety mechanisms such as fairness audits, drift monitoring, and bias detection are largely absent, leaving pipelines vulnerable to regulatory and ethical risks~\cite{farooq2024survey}. Multimodal feature engineering, integrating text, images, or graph data, remains underexplored despite its importance for enterprise workflows.

Advancing this area requires domain-adaptive benchmarks that assess feature quality independently of task performance, cross-modal pipelines capable of integrating heterogeneous sources, and safety-aware designs that embed bias audits and drift detection. Combining RL-based optimization with interpretable reasoning offers a promising path toward adaptive, transparent, and accountable feature engineering systems.

\subsection{Model Building and Selection}

Agents in this stage automate model training, selection, and evaluation. This involves choosing appropriate algorithms, preparing training data, tuning hyperparameters, and assessing performance. These systems aim to optimize predictive accuracy while ensuring fairness, robustness, reproducibility, and interpretability.

\subsubsection{\textbf{Conventional Supervised Training and Selection}}
Model building and selection in the data science lifecycle traditionally involved manually choosing algorithms, tuning hyperparameters, and evaluating models on fixed datasets to maximize predictive accuracy and efficiency. LLM-based agents transform these steps into adaptive and dynamic processes, reasoning over algorithmic choices, orchestrating external tools, and refining strategies iteratively \cite{zhang2024benchmarking,farooq2024survey}.

Agentic DS systems decompose model building into subtasks such as selecting candidate algorithms, performing hyperparameter search, and validating results, while iteratively refining strategies based on evaluation feedback. Four representative systems illustrate different optimization priorities. AutoKaggle \cite{li2024autokaggle} excels at orchestrating collaborative multi-agent workflows and achieves high validation success rates in simulated Kaggle competition settings. DS-Agent \cite{guo2024ds} leverages case-based reasoning to reuse past successful pipelines and iteratively construct and refine models, offering stronger adaptability across heterogeneous datasets. AutoML-GPT \cite{zhang2023automl} prioritizes end-to-end automation of the full training loop, covering preprocessing, architecture selection, and hyperparameter tuning across diverse datasets. AgentTuning \cite{zeng2023agenttuning} emphasizes lightweight instruction tuning, matching GPT-3.5 performance without sacrificing general NLP competence and providing a stable, efficient alternative to more resource-intensive methods.  This diversity reflects distinct design goals: competition robustness, cross-domain transfer, comprehensive automation, and computational efficiency.

Adaptability is further enhanced through parameter-efficient fine-tuning (PEFT) methods such as LoRA and QLoRA, which minimize computational cost while enabling domain-specific specialization \cite{parthasarathy2024ultimate}. RL methods, including Proximal Policy Optimization (PPO), are used to optimize multi-step reasoning and preference alignment for complex modeling tasks \cite{farooq2024survey}. Tool-augmented training, as introduced in Toolformer, enables agents to decide when and how to call APIs for tasks such as hyperparameter search or model ensembling \cite{schick2023toolformer}. Hybrid architectures combine general-purpose LLM planners with domain-specific models; for example, BloombergGPT \cite{wu2023bloomberggpt} demonstrates strong performance in finance-specific modeling while retaining broader NLP competence, highlighting the promise of vertical-domain agents.

Current DS agents operate exclusively on static, offline datasets. None of the 45 surveyed systems demonstrate continuous adaptation capabilities such as incremental retraining, automated drift detection, or dynamic hyperparameter adjustment in response to evolving data distributions. This gap is particularly limiting for domains where models must adapt to shifting patterns, such as recommender systems tracking changing user preferences or financial models responding to market volatility. Three critical limitations underlie this constraint. First, reproducibility is undermined by stochastic search strategies and LLM reasoning, hindering consistent results. Second, domain transfer remains weak, with models often failing outside their training contexts. Third, existing evaluation benchmarks focus on static datasets rather than dynamic scenarios. Closing these gaps will require explainable model search workflows, rigorous provenance logging, and cross-domain evaluation benchmarks that assess adaptability alongside accuracy.

% \textcolor{black}{Beyond static datasets, LLM-based DS agents could be extended to domains requiring continuous adaptation. For instance, in personalized recommender systems, agents could query external APIs to retrieve real-time user activity or contextual information; in finance, they could stream market data and sentiment analysis results for time-sensitive modeling. Such capabilities remain underutilized, as most current agents operate on offline data. Future DS agents could also automate full model lifecycle management, including incremental retraining, drift detection, model refreshing with new data, and continuous hyperparameter optimization. These adaptive pipelines would maintain relevance in dynamic environments, a direction largely unexplored in current benchmarks.}

% Despite advances, three limitations remain. First, reproducibility is undermined by stochastic search strategies and LLM reasoning, hindering consistent results. Second, domain transfer is weak, with models often failing outside their training context. Third, none of the 45 surveyed systems support continuous adaptation, including incremental retraining, drift detection, or dynamic hyperparameter adjustment. Closing these gaps requires explainable model search, rigorous logging, and cross-domain benchmarks tailored for DS agents.

\subsubsection{\textbf{Reinforcement Learning and Preference Optimization for Agent Training}} 

While RL and preference optimization have proven valuable for training general-purpose agents on multi-step reasoning tasks, their adoption in DS agents remains limited. This section examines existing RL applications in DS agents, explains why adoption remains minimal despite theoretical promise, and identifies underexplored methods that could improve multi-step planning, tool orchestration, and alignment with human preferences.

% \vspace{-0.5\baselineskip}
\paragraph{\textbf{RL and Preference-Based Alignment}}
While supervised fine-tuning (SFT) provides strong initialization, it often fails on multi-step data science workflows that require planning, tool orchestration, and error recovery \cite{zeng2024agenttuning,laleh2024survey}. RL addresses these gaps by optimizing policies through reward signals \cite{ouyang2022training}, while RL from Human Feedback (RLHF) aligns model behavior with user preferences using reward models and Proximal Policy Optimization (PPO) \cite{schulman2017ppo}. Applied to DS pipelines, these methods hold promise for improving reasoning across data cleaning, feature engineering, modeling, and reporting, though adoption remains limited. To reduce the cost of RLHF, lightweight alignment methods have emerged. Direct Preference Optimization (DPO) fine-tunes models directly on preference pairs without explicit reward models \cite{rafailov2023direct}, while Group-Relative Policy Optimization (GRPO) extends this to group-wise rankings \cite{shao2024deepseekmath}. AgentTuning \cite{zeng2024agenttuning} shows that high-quality demonstrations can induce robust planning and tool-use skills for EDA, feature engineering, and evaluation without full RLHF overhead. Current data science agents do not yet employ preference-based alignment. While systems such as AgentTuning illustrate this approach in broader agentic settings, methods like DPO and GRPO remain unexplored for DS workflows despite their potential to align feature-engineering suggestions, visualizations, explanations, and summaries with user intent, support model ranking and next-best-action recommendations, and refine domain-specific outputs. This highlights alignment as a key open challenge for future DS agents.

% \vspace{-0.5\baselineskip}
\paragraph{\textbf{RL for Planning and Sequential Tool Use}}
A defining capability of DS agents is the ability to plan multi-step workflows and chain tools to achieve high-level goals. Traditional prompt-based approaches often fail when faced with issues such as faulty SQL queries, misleading charts, or unexpected data distributions \cite{zeng2024agenttuning}. RL reframes data science as a sequential decision-making process, allowing agents to learn tool-use strategies through trial and error, guided by reward signals and long-term planning \cite{laleh2024survey}. Hierarchical planning supports this process by decomposing large tasks into sub-goals and incorporating correction cycles. In data preparation and feature engineering, tasks like handling missing values or generating features can be managed as subtasks coordinated by planner–executor modules. During model training and evaluation, frameworks such as Plan-and-Act \cite{erdogan2025planandact} and LLaMAR \cite{nayak2025llamar} enable iterative plan–act–verify cycles, while in deployment and monitoring, hierarchical planning facilitates error detection, retraining, and adaptation. Orchestration frameworks like LangGraph \cite{langgraph2024} and AutoGPT \cite{yang2023autogpt} extend these ideas by chaining reasoning steps and revising intermediate outputs based on feedback.

Several RL-based systems demonstrate these capabilities in practice. SWE-RL \cite{wei2025swerl} trains a 70B-parameter model to resolve GitHub issues through multi-step reasoning and external tool use, mirroring debugging stages of data pipelines. WebRL \cite{qi2025webrl} enables models to act as web agents, acquiring and preparing data by navigating diverse sources and adapting to changing conditions. Toolformer \cite{schick2023toolformer} fine-tunes models to decide when to call APIs, supporting exploratory data analysis and interpretation. Deep Research \cite{wei2025browsecomp} shows how RL can improve information retrieval by training agents to query search engines, evaluate page relevance, and refine strategies. These systems align with knowledge discovery and continuous learning, where agents must iteratively gather, filter, and interpret information. Recent foundation models such as GPT-4o \cite{achiam2023gpt}, Gemini \cite{gemini2.5}, Claude \cite{claude}, DeepSeek-R1 \cite{deepseekai2025deepseekr1incentivizingreasoningcapability}, Qwen \cite{qwen2025qwen25technicalreport}, LLaMA \cite{llama4}, and Magistral \cite{magistral} already demonstrate sequential tool use, signaling a broader shift toward agentic systems that can autonomously plan, act, and adapt across the data-science lifecycle.
\vspace{-0.5\baselineskip}
\paragraph{\textbf{Multi-Agent Reinforcement Learning (MARL)}}
As data science tasks grow in scale and complexity, single agents often struggle to cover the full range of required skills. MARL enables multiple agents to divide tasks, exchange intermediate outputs, and coordinate actions under shared or partially aligned reward signals. Systems like AutoGen \cite{wu2023autogenenablingnextgenllm} show how role-specialized agents (e.g., planner, coder, critic) collaborate to solve SWE-bench tasks more efficiently. In data-science workflows, this translates to one agent extracting tables, another generating visualization code, and a third validating insights, all jointly optimized for correctness and fidelity. MARL can also support multi-user settings by adapting to heterogeneous goals \cite{zhang2024benchmarking}, while maintaining shared memory, cached computations, and reusable plots to improve efficiency in long-running analyses.

Despite these opportunities, RL adoption in DS agents remains minimal. RLHF is resource-intensive, requiring large preference datasets, reward models, and iterative PPO optimization \cite{schulman2017ppo,farooq2024survey}, which becomes costly when actions involve code execution or API calls. Sample inefficiency and sparse rewards hinder multi-step learning, while goals like interpretability or insight quality are difficult to formalize \cite{wei2025swe}, leading agents to exploit proxy metrics \cite{wu2023reasoning}. Policies trained on curated data also struggle to generalize to new schemas or APIs \cite{laleh2024survey}. RL-based optimization is largely missing from current data science agents, with no systems yet demonstrating multi-stage credit assignment or stable deployment in production workflows. Addressing these gaps will require hybrid strategies that combine RL with expert imitation, modular policy architectures, curriculum learning, and human-in-the-loop preference shaping \cite{zeng2024agenttuning,shao2024deepseekmath}.

\subsection{Interpretation and Explanation}

Interpretability and explainability are essential for building trust in agentic data science workflows, where automated decisions span multiple pipeline stages. Agents must justify not only final predictions but also intermediate choices such as why features were selected, why a model architecture was chosen, or how hyperparameters were tuned in ways that are transparent, faithful, and actionable. Without this capability, systems risk producing outputs that are statistically sound but opaque and difficult to validate or debug. 

Current DS agents with explanation capabilities focus primarily on exploratory data analysis and visualization-driven storytelling rather than model-level interpretability. Jupybara~\cite{wang2025jupybara} leverages multi-agent prompting to generate coherent narratives and interactive EDA insights. XMODE~\cite{nooralahzadeh2024explainable} supports explainable multimodal exploration by decomposing queries into subtasks such as SQL generation and image analysis. Data Formulator 2~\cite{wang2025data} enables mixed-initiative visualization with AI-driven transformations, while InsightBench~\cite{sahu2024insightbench} evaluates agents on their ability to communicate actionable business insights, explicitly scoring explanation quality. These systems demonstrate value in explaining data patterns but do not incorporate mechanisms for model interpretability such as feature attribution, decision-path tracing, or mechanistic analysis of learned representations.

Several explanation methods could bridge this gap. Classical techniques like SHAP~\cite{lundberg2019explainable} and LIME~\cite{ribeiro2016should} provide feature-level attributions for model predictions. LLM-specific methods include (i) prompt-based self-explanations such as Chain-of-Thought~\cite{wei2022chain} and self-consistency~\cite{zhao2024explainability}, which generate intermediate reasoning steps, (ii) attribution on embeddings by adapting SHAP/LIME to token-level representations or analyzing attention patterns~\cite{yu2025survey}, and (iii) mechanistic interpretability, which isolates neurons, attention heads, or functional subnetworks that influence decisions~\cite{piccialli2025agentai}. Applied to DS workflows, these approaches could explain model selection, feature engineering, and reporting choices in ways that support debugging, validation, and regulatory compliance.

Despite progress, explanation in agentic pipelines remains limited. Post-hoc methods often lack faithfulness, yielding rationalizations that fail under distribution shift, and evaluation is fragmented, with no benchmarks assessing explanation quality across full DS lifecycles. High computational cost, privacy and provenance constraints, and the absence of task-specific metrics further restrict scalability and practical utility. Future progress requires standardized benchmarks for faithfulness, integration of interpretability into decision-tracking pipelines, and lightweight methods that ensure explanations are both transparent and efficient, especially in high-stakes domains where auditability and regulatory compliance are essential.

\subsection{Deployment and Monitoring}

Deployment and monitoring are critical for ensuring that models transition from development to production reliably while maintaining performance over time \cite{siddiqui2023comprehensive}. In classical ML pipelines, this involves CI/CD automation, observability, and governance. Agentic systems extend these workflows by enabling adaptive monitoring, automated remediation, and compliance-aware decision-making. For example, a retail forecasting agent can be deployed via canary rollout on a subset of transactions, with automatic rollback if accuracy degrades. Once live, the agent can track seasonal drift and trigger retraining, weight adjustment, or reversion to a stable version when stability declines. 

Only a limited number of DS agents currently provide deployment capabilities, and most pipelines still depend on manual or semi-automated steps. AutoML-Agent~\cite{trirat2024automl} integrates deployment as the final stage of its pipeline, generating deployment-ready models and inference endpoints with minimal intervention. DS-Agent~\cite{guo2024ds} supports deployment in low-resource settings by reusing prior case solutions to generate executable training and inference code, reducing retraining overhead and accelerating operationalization. In production environments, agentic systems can leverage containerization (e.g., Docker), orchestration (e.g., Kubernetes), and CI/CD frameworks (e.g., Jenkins, GitLab CI) for automated build, test, and release cycles~\cite{ugwueze2024continuous,siddiqui2023comprehensive}. Observability stacks such as Prometheus and Grafana, combined with telemetry frameworks like OpenTelemetry, provide real-time monitoring through dynamic dashboards. Agents can also prepare Airflow DAGs, schedule jobs, and detect ML-specific issues such as concept drift or silent failures by comparing live predictions with baselines.

While tool integration shows promise, current systems lack seamless automation of monitoring and recovery. Drift detection is often reactive rather than proactive, and rollback or retraining pipelines require extensive manual oversight. Even when orchestration tools are in place, automated workflows rarely integrate governance requirements such as policy-as-code, audit logging, and human oversight checkpoints~\cite{raza2025trism}. Embedding these safeguards directly into agent workflows will be essential for achieving hands-free, production-grade deployment. Future research should focus on building agents that can continuously monitor performance, trigger context-aware remediation, and balance reliability with compliance and transparency in enterprise settings.

\definecolor{headerblue}{RGB}{226,239,255}
\definecolor{rowgray}{gray}{0.96}

\begin{table}[!htbp]
\scriptsize
\centering
\caption{Stage-level synthesis of agentic capabilities across the data science lifecycle, including gaps in safety mechanisms.}
\label{tab:lifecycle-synthesis}
\setlength{\tabcolsep}{3pt}
\rowcolors{2}{rowgray}{white}
\begin{tabular}{|p{2.4cm}|p{5.0cm}|p{5.4cm}|p{2cm}|}
\hline
\rowcolor{headerblue}
\textbf{Lifecycle Stage} & \textbf{Strengths (Progress Made)} & \textbf{Struggles / Gaps} & \textbf{Coverage} \\
\hline
Business Understanding \& Data Acquisition\textbf{ (S1)} &
Goal translation; schema parsing; anomaly detection; structured acquisition &
Ambiguous intent; fragile orchestration; weak privacy, fairness, and compliance safeguards &
\textbf{Limited} (early prototypes) \\
\hline
Exploratory Data Analysis \& Visualization \textbf{(S2)} &
Descriptive stats; chart generation; narrative reporting; multimodal reasoning; interactive analysis &
Faithfulness errors; scaling issues; domain misinterpretation; hallucinations; limited bias/safety checks &
\textbf{Well covered} (most agents) \\
\hline
Feature Engineering \textbf{(S3)} &
Automated transformations; semantic feature creation; feature selection; explainability &
Cross-domain fragility; poor non-tabular support; weak drift monitoring; safety guardrails missing &
\textbf{Moderate} (several systems) \\
\hline
Model Building \& Selection \textbf{(S4)} &
Algorithm choice; hyperparameter tuning; adaptive refinement &
Fairness gaps; reproducibility issues; unstable evaluation; safety and bias audits rarely integrated &
\textbf{Partial} (many systems, shallow) \\
\hline
Interpretation \& Explanation \textbf{(S5)} &
Visual explanations; narrative reporting; transparency; mixed-initiative support &
Brittle outputs; hallucinations; poor uncertainty calibration; lack of standardized safety validation &
\textbf{Partial} (growing, inconsistent) \\
\hline
Deployment \& Monitoring \textbf{(S6)} &
Drift detection; audit logging; governance awareness &
Rare in practice; weak security; absent continuous monitoring; limited trust and compliance mechanisms &
\textbf{Very limited} (least represented) \\
\hline
\end{tabular}
\label{capab}
\end{table}

\noindent \textbf{Cross-Cutting Oversight.} Beyond these six stages,
Human-in-the-Loop (HITL) oversight remains a cross-cutting component across them. By embedding human feedback at critical points such as exploratory data analysis, feature selection, and model interpretation, HITL improves transparency, adaptability, and reliability in high-stakes or ambiguous scenarios \cite{zhang2024benchmarking}. Systems like Agentpoirot \cite{sahu2024insightbench} and DatawiseAgent \cite{you2025datawiseagent} demonstrate that user interventions can enhance explanation quality, auditability, and decision safety, while confirmation before sensitive actions (e.g., retraining, data deletion) mitigates risks.

Table~\ref{tab:lifecycle-synthesis} provides a stage-level synthesis of the capabilities and limitations of current data science agents. For each stage of the lifecycle, it outlines where progress has been made (e.g., chart generation, feature selection, algorithm tuning), where struggles and gaps persist (e.g., fairness, reproducibility, safety validation, deployment monitoring), and the extent of coverage across surveyed systems.

\section{Evaluation and Benchmarking}
\label{sec:evaluation}

Evaluating agentic data science systems poses unique challenges, arising from their multi-step autonomy, multimodal reasoning, tool invocation, and variable workflows. Traditional single-metric evaluation (e.g., accuracy or latency) is insufficient to capture these agents' full capabilities and limitations.

\subsection{Core Evaluation Dimensions}
\label{sec:eval-dimensions}

Evaluating agentic data science systems requires moving beyond traditional metrics. Because these systems operate through multi step workflows, multimodal reasoning, and dynamic tool use, evaluation must capture their functional, ethical, and operational quality~\cite{zhang2024benchmarking, zhang2024multitrust}. Five core dimensions are central: (i) Task Effectiveness, measuring whether the agent completes tasks correctly, fully, and efficiently; (ii) Trustworthiness, reflecting fairness, privacy, and robustness of outputs; (iii) Explainability, assessing whether reasoning and decisions are communicated transparently; (iv) Efficiency, considering computational cost and latency, tool usage, and human effort; and (v) User Satisfaction, capturing user trust, preference, and overall interaction quality. Section~\ref{sec:eval-metrics} expands these into measurable criteria, including robustness and modality specific quality.

As these dimensions often involve trade offs, such as speed versus interpretability or privacy versus accuracy, multi objective evaluation is essential. \citet{zhao2024explainability} recommend reporting faithfulness, completeness, and stability as foundational criteria alongside functional metrics. Furthermore, Chain-of-Thought (CoT) rationales, increasingly adopted in datasets such as GSM8K and BIG-Bench~\cite{wei2022chain}, enable intermediate-step grading, which supports diagnosing reasoning fidelity rather than evaluating only the final results.

\subsection{Functional vs.\ Process-Centric Evaluation}
\label{sec:eval-process-centric}

Traditional evaluations often emphasize functional success, that is whether the agent produces the correct output. While this captures end task performance through metrics such as accuracy, F1 score, or code execution success, it provides little insight into how results are derived. For agentic systems operating through multi step workflows, this limitation is significant, since a correct final result may mask critical errors or unsafe practices in intermediate steps~\cite{zhang2024benchmarking, gu2024blade}.

Process centric evaluation addresses this gap by examining intermediate reasoning, tool usage, error propagation, and adherence to valid workflows. For example, an agent might output the correct final result while silently corrupting internal states such as in place dataframe edits, or by applying incorrect control variables, which undermines the reliability and reproducibility of analysis~\cite{zhang2024benchmarking, gu2024blade}. Benchmarks such as DSEval and InsightBench therefore advocate dual evaluation strategies that score both final accuracy and the integrity of reasoning, tool orchestration, and session state, ensuring comprehensive and trustworthy assessment of agent performance.

\subsection{Benchmark Datasets and Frameworks}
\label{sec:eval-benchmarks}

Several benchmarks have been introduced to evaluate agentic data science systems across tasks such as database querying (Spider 2.0, Spider2-V), pipeline orchestration (ELT-Bench, BrowseComp), exploratory analysis and visualization (InsightBench, MatplotBench), and statistical reasoning (BLADE). General purpose frameworks such as AgentBench and TauBench assess multi tool workflows and robustness. In addition, platforms such as OpenML provide a growing suite of standardized benchmarks that support reproducible experimentation across diverse machine learning and data science tasks~\cite{vanschoren2014openml}. Table~\ref{tab:eval-benchmarks} compares these benchmarks, highlighting their objectives, task modalities, and evaluation strategies.

\definecolor{headerblue}{RGB}{226,239,255}
\definecolor{rowgray}{gray}{0.96}

\begin{table}[htbp]
\scriptsize
\centering
\caption{Comparison of Key Benchmarks for Data Science Agents. \footnotesize Stage: S\# = lifecycle, NL = non-lifecycle}
\label{tab:eval-benchmarks}
\setlength{\tabcolsep}{3pt}
\rowcolors{2}{rowgray}{white}
\begin{tabular}{|p{2.5cm}|p{0.9cm}|p{3.9cm}|p{3.4cm}|p{3.9cm}|}
\hline
\rowcolor{headerblue}
\textbf{Benchmark} & \textbf{Stage} & \textbf{Task Domains / Tooling} & \textbf{Dataset Size} & \textbf{Evaluation} \\
\hline

ELT-Bench \cite{jin2025elt} % [7 Apr 2025]
& S1, S3 & Airbyte, Terraform, dbt, Snowflake, S3, PostgreSQL, MongoDB & 100 pipelines, 835 tables, 203 data models & Execution-based: SRDEL, SRDT \\

IDA-Bench \cite{li2025ida} % [23 May 2025]
& S2-S4 & Python sandbox, pandas, NumPy, sklearn, csv & 25 notebooks, 209 insights & Exec-based vs human baseline (accuracy, MSE) \\

TimeSeriesGym \cite{cai2025timeseriesgym} % [19 May 2025]
& S2-S4 & Time series, Python, PyTorch, GitHub repos & 34 challenges, 23 sources, 8 problem types & Exec-based metrics, LLM judge \\

DataSciBench \cite{zhang2025datascibench} % [19 Feb 2025]
& S2-S4 & Python sandbox, plotting libs, sklearn & 222 prompts, 519 cases, 6 tasks & TFC, Exec-based scoring, VLM-judge \\

LLM4DS \cite{nascimento2024llm4ds} % [16 Nov 2024]
& S2 & Python, Matplotlib, Stratascratch & 100 coding problems & Execution-based success rate \\

Spider 2.0 \cite{lei2024spider} % [12 Nov 2024]
& S2-S3 & BigQuery, Snowflake, SQLite, DuckDB & 632 tasks, 213 DBs & Execution-based, success rate, accuracy \\

DA-Code \cite{huang2024code} % [9 Oct 2024]
& S2-S4 & Python, SQL, Bash & 500 tasks (DW 100, ML 100, EDA 300) & Exec-based, table/chart match, ML metrics \\

DSBench \cite{jing2024dsbench} % [12 Sep 2024]
& S2-S4 & Excel, tables, Python, images & 466 analysis, 74 modeling tasks & Exec-based, accuracy, performance gap \\

BLADE \cite{gu2024blade} % [19 Aug 2024]
& S2-S4 & Python, pandas, statsmodels & 12 datasets / 536 decisions & LM-assisted decision matching \\

Spider2-V \cite{cao2024spider2} % [15 Jul 2024]
& S2-S3 & BigQuery, Snowflake, Airbyte, GUI apps & 494 tasks; 170 setups; 151 scripts & Exec-based state checks, validators \\

InsightBench \cite{sahu2024insightbench} % [8 Jul 2024]
& S2-S5 & Python, pandas, business ops data & 100 datasets (500 rows each) & LLaMA-3-Eval, ROUGE-1 \\

tau-bench \cite{yao2024tau} % [17 Jun 2024]
& S1, S6 & Python API tools, LM user sim & 165 tasks (115 retail, 50 airline) & Exec-based DB match, pass@k \\

Tapilot-Crossing \cite{li2024tapilot} % [8 Mar 2024]
& S2-S4 & Python, pandas, numpy, four modes & 1024 interactions, 1176 intents & Execution-based accuracy (Acc, AccR) \\

DSEval \cite{zhang2024benchmarking} % [27 Feb 2024]
& S2-S5 & Pandas, NumPy, sklearn, Jupyter & 294 sets, 825 problems & Exec-based, pass rate \\

MatPlotBench \cite{yang2024matplotagent} % [18 Feb 2024]
& S2 & Python Matplotlib, OriginLab, Bokeh & 100 test cases & GPT-4V score \\

InfiAgent-DABench \cite{hu2024infiagent} % [10 Jan 2024]
& S2-S4 & Python, pandas, scikit-learn, ReAct & 257 questions, 52 CSVs & Exact-match accuracy \\

SWE-bench \cite{jimenez2023swe} % [10 Oct 2023]
& NL & GitHub repos, patch generation, unit tests & 2294 tasks, 12 repos & Execution-based patch success \\

STRUC-BENCH \cite{tang2023struc} % [16 Sep 2023]
& S3 & Raw text, HTML, LaTeX, FormatCoT & 3400/728 raw, 5300/500 LaTeX, 5400/499 HTML & P-Score, H-Score, BLEU, ROUGE \\

AgentBench \cite{liu2023agentbench} % [7 Aug 2023]
& NL & Bash, SQL, APIs, ALFWorld, WebShop, Mind2Web & 1360 tasks (269 dev / 1091 test) & Env success rate, F1, reward, OA score \\

ARCADE \cite{yin2022natural} % [19 Dec 2022]
& S2-S3 & pandas, Jupyter notebooks & 1082 problems, 136 notebooks & Exec-based fuzzy output matching \\
\hline
\end{tabular}
\end{table}

\subsection{Evaluation Metrics and Key Performance Indicators (KPIs)}
\label{sec:eval-metrics}

Evaluating agentic data science systems requires a diverse set of metrics that go beyond traditional accuracy or latency, reflecting multi-step workflows, multimodal reasoning, and tool orchestration. These metrics can be broadly grouped into five categories: \emph{(i) task correctness}, \emph{(ii) pipeline-level success}, \emph{(iii) robustness and reliability}, \emph{(iv) quality of outputs (visualizations, data, and models)}, and \emph{(v) efficiency and usability}. Below we summarize key measures within each category.
% \vspace{-0.5\baselineskip}
\paragraph{\textbf{Task Correctness.}} Standard accuracy and F1-score remain fundamental for classification, extraction, and multiple-choice tasks~\cite{naidu2023review, gu2024blade, hu2024infiagent}. Execution-based metrics such as Execution Accuracy (EX) check whether generated SQL queries return correct results irrespective of irrelevant columns or row order~\cite{lei2024spider}. Functional correctness and fuzzy output matching check semantic equivalence of execution outputs, accommodating minor formatting variations~\cite{yin2022natural}. For code similarity, metrics such as Code Similarity Equivalence (CSE), Accuracy for Similarity Evaluation (AccSE), and hybrid embedding-based similarity approaches capture partial correctness beyond binary execution success~\cite{li2024tapilot}. Accuracy variants like AccR extend correctness checks to ensure correct usage of private libraries~\cite{li2024tapilot}.
% \vspace{-0.5\baselineskip}
\paragraph{\textbf{Pipeline-Level Success.}} Metrics such as \emph{Success Rate (SR)} compute the proportion of tasks fully completed~\cite{lei2024spider, nascimento2024llm4ds, cao2024spider2}. Specialized forms include SRDEL and SRDT for ELT pipelines, capturing success in extraction/loading and transformation phases respectively~\cite{jin2025elt}. Completion Rate (CR) provides a normalized 0–1 score over multi-step tasks by scoring each step as missing, non-compliant, or fully compliant~\cite{zhang2025datascibench, huang2024code}. Other indicators include Valid Submission (\%), which measures syntactic and execution-level correctness of submission files, and Reasonable Submission (\%), which applies a quality threshold such as leaderboard ranking~\cite{li2025ida, cai2025timeseriesgym}. For holistic evaluation, some benchmarks use Overall Score, combining table match, chart match, and ML performance into a unified metric~\cite{huang2024code}.
% \vspace{-0.5\baselineskip}
\paragraph{\textbf{Robustness and Reliability.}} Metrics like pass@\emph{k}, pass$^{k}$, and pass$\hat{k}$ evaluate robustness under multiple independent runs, indicating the probability of success in at least one run or all runs across trials~\cite{jin2025elt,yao2024tau, yin2022natural}. Variants of Pass Rate further capture robustness by comparing strict validation against relaxed conditions, including settings that account for error propagation, ignore intactness violations, or disregard presentation errors~\cite{zhang2024benchmarking}. Related measures include Relative Performance Gap (RPG), which normalizes performance against human baselines, and Coverage@k or Adapted F1, which combine correctness and diversity across sampled outputs~\cite{jing2024dsbench, gu2024blade}.
% \vspace{-0.5\baselineskip}
\paragraph{\textbf{Output Quality (Data, Visualization, and Models).}} For visualization tasks, Visualization Similarity Score and Automatic Plot Quality Score compare generated charts against ground truth using vision–language models such as GPT-4V~\cite{yang2024matplotagent, nascimento2024llm4ds}. MatplotBench validates these with human correlation scores (Pearson \(r>0.8\))~\cite{yang2024matplotagent}. Additional quality dimensions include VLM-as-a-judge scores (0–5 rubric for axes, legends, annotations), and structural checks such as \emph{Plot Validity (F2)} and \emph{Visualization Completeness (F4)}~\cite{zhang2025datascibench}. Data-oriented metrics include Data Quality (F1) and Data Accuracy (F3) via error thresholds (e.g., MSE). For modeling tasks, Model Accuracy (F5) and normalized ML scores capture predictive performance~\cite{zhang2025datascibench, huang2024code}. Insight generation quality is assessed via LLM-based metrics such as LLaMA-3-Eval, which correlates well with human judgment, and ROUGE-1 for text similarity~\cite{sahu2024insightbench}.
% \vspace{-0.5\baselineskip}
\paragraph{\textbf{Efficiency and Usability.}} Runtime efficiency is measured by median execution time for accepted runs~\cite{nascimento2024llm4ds}. The Executable-Code Ratio quantifies robustness by counting the proportion of generated scripts that run without errors~\cite{huang2024code}. For dynamic evaluation, some frameworks adopt human-centered scores or reward-based proxies, such as Average Reward (pass$^1$) in interactive multi-turn tasks~\cite{yao2024tau}.

% \subsection{Human-in-the-Loop and Subjective Evaluation}
% \label{sec:human-eval}

% While automated metrics offer scalability and consistency, many critical aspects of agentic performance—such as trust, usability, and ethical alignment—require human judgment. Human-in-the-loop evaluations typically involve structured user studies to measure satisfaction, trust, and clarity, as well as expert audits that assess process quality, tool appropriateness, and compliance with domain-specific regulations, particularly in high-stakes settings like healthcare or finance. Preference-based evaluation methods, including prompt judging and pairwise comparisons, are commonly employed in reinforcement learning with human feedback (RLHF) or direct preference optimization (DPO) to capture nuanced dimensions such as helpfulness and correctness. In addition, recent benchmarks such as BLADE~\cite{gu2024blade} and InsightBench~\cite{sahu2024insightbench} advocate process-alignment scoring, where agent reasoning traces are compared to expert workflows for interpretability and reliability. Finally, explanation quality is often rated by human judges on coherence, faithfulness, and helpfulness, providing essential insights for building transparent and user-aligned systems. Although resource-intensive, these evaluations remain indispensable for complementing automated metrics and ensuring that agentic systems meet human-centric standards of safety and trust.

\subsection{Human-in-the-Loop and Subjective Evaluation}
\label{sec:human-eval}

While automated metrics provide scalability and consistency, many critical aspects of agentic performance, such as trust, usability, and ethical alignment, require human judgment. Human-in-the-loop evaluations typically involve structured user studies that measure satisfaction, trust, and clarity. They also include expert audits that assess process quality, tool appropriateness, and compliance with domain-specific regulations, particularly in high-stakes settings such as healthcare or finance. Preference-based evaluation methods, including prompt judging and pairwise comparisons, are widely used in RLHF or DPO to capture nuanced dimensions such as correctness and helpfulness. Recent benchmarks such as BLADE~\cite{gu2024blade} and InsightBench~\cite{sahu2024insightbench} further advocate process-alignment scoring, where agent reasoning traces are compared against expert workflows to evaluate interpretability and reliability. In addition, human judges are often needed to assess visualization quality, alongside explanation quality, rating outputs on coherence, faithfulness, and usefulness.  Despite their cost, these evaluations remain indispensable for complementing automated metrics and ensuring that agentic systems meet human-centric standards of safety and trust.

\subsection{Current Gaps in Evaluation}
\label{sec:evaluation-challenges}

Despite significant progress, evaluating agentic data science systems remains challenging due to the complexity of multi-step, multimodal workflows. Current evaluation practices lack unified standards, resulting in fragmented metrics that often fail to capture process fidelity or robustness. For instance, benchmarks such as DSEval reveal that up to 27\% of failures stem from silent state integrity violations (e.g., in-place dataframe mutations) even when final outputs are correct~\cite{zhang2024benchmarking}. Similarly, statistical reasoning errors, such as misidentified control variables or invalid transformations highlighted by BLADE~\cite{gu2024blade}, are rarely detected by syntax-based or crash-based evaluations. These gaps highlight the need for more granular, process-aware metrics that go beyond end-task correctness.

Beyond fragmented metrics, current benchmarks also expose recurring failure modes that highlight limitations of existing evaluation practices. For example, poor multi-language orchestration leads agents to hallucinate filenames or crash during cross-language execution (DA-Code~\cite{huang2024code}). Performance also drops sharply across pipeline stages, with ELT-Bench reporting 57\% success for data loading compared to only 3.9\% for transformation~\cite{jin2025elt}. Similarly, Spider 2.0 and Spider2-V highlight fragility in handling large schemas and interface-driven workflows~\cite{lei2024spider,cao2024spider2}. These issues, compounded by the high costs of human-in-the-loop evaluation, limit the scalability, reliability, and interpretability of agentic systems. Addressing these limitations calls for standardized, robustness-aware frameworks that integrate process-level diagnostics with modality-specific quality checks.

\section{Open Challenges and Future Directions}

\label{sec:outlook}

Despite rapid progress, our analysis of 45 agentic data science systems reveals that significant challenges remain before these systems can achieve reliable deployment in real-world settings. Current approaches often fall short in robustness, trust, safety, and scalability, with gaps in benchmarking and evaluation further limiting systematic progress. Addressing these challenges requires not only technical advances in learning, reasoning, and alignment, but also careful consideration of compliance, human oversight, and socio-economic impact. The following subsections highlight key open problems and outline future research directions (see Tab.  \ref{tab:challenges_future}).

\definecolor{headerblue}{RGB}{226,239,255}
\definecolor{rowgray}{gray}{0.96}
\setlength{\arrayrulewidth}{0.4pt} % thin but visible borders
% \arrayrulecolor{black} % (optional) ensure black rules

% --- TABLE ---
\begin{table}[htbp]
\scriptsize
\centering
\caption{Key Challenges and Future Directions for Data Science Agents}
\label{tab:challenges_future}
\setlength{\tabcolsep}{3pt}
\renewcommand{\arraystretch}{1.12}

\rowcolors{2}{rowgray}{white} % start alternating colors on 2nd row of the body
\begin{tabular}{|p{3.8cm}|p{5.0cm}|p{6.2cm}|}
\hline
\rowcolor{headerblue}
\textbf{Challenge} & \textbf{Description} & \textbf{Future Direction} \\
\hline

\textit{Prompt Ambiguity} &
Data science problems are often under-specified or ill-defined. &
Equip agents to ask clarifying questions and disambiguate user intent before execution. \\
\hline

\textit{Memory \& Long-Term Planning} &
Poor retention of context across long workflows leads to broken reasoning. &
Use memory modules (e.g., RAG, MemoryBank), long-context models, multi-agent decomposition, async execution, and fault-tolerant recovery. \\
\hline

\textit{Security \& Privacy} &
Weak defenses against data leaks and adversarial inputs. &
Enforce privacy, adversarial testing, secure sandboxes, and governance protocols. \\
\hline

\textit{Regulatory Compliance} &
Legal constraints in sensitive domains (finance, healthcare). &
Involve legal/compliance teams early; document decisions; keep humans in the loop for high-stakes tasks. \\
\hline

\textit{Trust \& Transparency} &
Lack of explainability limits adoption. &
Design interpretable and auditable workflows with fairness, traceability, and error provenance. \\
\hline

\textit{Reliability \& Hallucination} &
Multi-step workflows can amplify small input errors, causing hallucinated code, results, or analyses. &
Combine prompt engineering with iterative debugging and unit testing; add neuro-symbolic checks (knowledge graphs, ontologies) and visualization-based reasoning for factual grounding. \\
\hline

\textit{Weak Feedback Loops \& Poor Alignment} &
Weak feedback channels from end-users will lead to poor alignment. &
Integrate feedback capture into UIs; use it for prompt refinement, preference tuning, and model updates. \\
\hline

%\textit{Control \& Oversight} &
%Fully autonomous agents make users uneasy; data scientists need guidance capabilities. &
%Let users preview plans and generated code, edit them pre-execution, and track each decision for accountability. \\
%\hline

\textit{Robustness \& Generalization} &
Agents may fail under unseen tasks or data distributions, leading to brittle performance. &
Train on diverse, representative datasets; use meta-learning and modular skills to enable domain adaptation and transfer learning. \\
\hline

\textit{Benchmark Gaps} &
Existing benchmarks test isolated subtasks only. &
Build end-to-end, multimodal lifecycle benchmarks reflecting real DS workflows. \\
\hline

\textit{Evaluation Gaps} &
Metrics focus on final output accuracy, ignoring reasoning quality. &
Use process-based evaluation and human-in-the-loop judgment to assess intermediate reasoning steps. \\
\hline

\textit{Scalability \& Efficiency} &
High compute and latency limit practical deployment. &
Use lightweight orchestration with smaller models; apply caching, batching, and adaptive computation. \\
\hline

%\textit{Human Escalation} &
%Agents lack clear escalation mechanisms for ambiguous cases. &
%Define human-in-the-loop workflows with confidence thresholds for escalation triggers. \\
%\hline

\textit{Ethical \& Socioeconomic Challenges} &
Fears of DS automation displacing human jobs. &
Position agents as co-pilots that augment, not replace, data scientists. \\

\hline

\textit{Multimodal Reasoning Gaps} &
Struggle with cross-modal grounding \& high costs &
Build efficient multimodal architectures and apply approaches like Visualization-of-Thought. \\

\hline

\end{tabular}
\end{table}

\subsection{Ambiguous Task Instruction}
Data science tasks are often under-specified, with users providing vague or incomplete prompts such as “analyze sales trends” or “generate insights.” In such cases, agents risk producing irrelevant or misleading outputs if they proceed without clarification. This ambiguity is particularly harmful in multi-stage workflows, where an early misinterpretation propagates downstream.

\noindent Future systems should be designed not merely as instruction-followers but as conversational partners that seek to understand the underlying business objective. These conversational partners must proactively ask clarifying questions, validate user intent, and refine task specifications before execution. For example, in response to ``analyze sales trends,'' an effective agent may ask the following clarification questions:

\begin{itemize}[leftmargin=*, nosep]
    \item 

\textbf{Scoping Questions:} \textit{``Over what time period should I analyze? Are you interested in a specific region or product line?''}

\item  \textbf{Comparison Questions:} \textit{``Should I compare this period's performance to the previous period, the same period last year, or against forecasted targets?''}

\item  \textbf{Goal-Oriented Questions:} \textit{``Is the goal to identify top-performing products, find areas for growth, or diagnose a recent dip in sales?''}
\end{itemize}
Recent frameworks demonstrate that this iterative intent clarification not only improves accuracy but also builds user trust by making the reasoning process transparent~\cite{zhang2024benchmarking, wu2023autogen}. Furthermore, this approach yields additional benefits by improving efficiency while preventing wasted computation on misguided analyses. Such dialogues can also educate users on how to formulate more precise and useful questions in the future.

%Data science tasks are often under-specified, with users providing vague or incomplete prompts such as “analyze sales trends” or “generate insights.” In such cases, agents risk producing irrelevant or misleading outputs if they proceed without clarification. This ambiguity is particularly harmful in multi-stage workflows, where an early misinterpretation propagates downstream. Future systems should be designed as conversational partners that proactively ask clarifying questions, validate user intent, and refine task specifications before execution. Recent frameworks demonstrate that iterative intent clarification not only improves accuracy but also builds user trust by making the reasoning process transparent~\cite{zhang2024benchmarking, wu2023autogen}.

% LLMs are sensitive to the input prompts. But if the data science problems are not well-defined, then the Agents may not be able to reliably perform the task. To address this, it is required to design the agent to be a conversational partner such that it asks clarifying questions in case of confusion.

\subsection{Limited Context Window} 
Data science workflows involve long-horizon reasoning, where later steps depend on earlier outputs. However, LLMs are constrained by limited context windows and often lose track of prior goals, leading to broken reasoning chains in multi-step pipelines. To address this, a more robust \textbf{planner-executor} paradigm is emerging as a powerful solution \cite{erdogan2025planandact}. Within this framework, a high-level \textbf{planner agent} strategically breaks down a complex goal into smaller verifiable subtasks. These granular tasks are then delegated to specialized \textbf{executor agents} that carry out each subtask. To preserve continuity over long workflows, the planner–executor approach can be enhanced with external vector databases or associative memory modules (e.g., LONGMEM or CAMELoT architectures) that can cache and retrieve information beyond the native context window \cite{wang2023augmenting,he2024camelot}. Other methods use context-management strategies, like summarizing or refreshing past steps, to prevent the model from getting overloaded, as seen in frameworks like Sculptor \cite{,li2025sculptor}. Recent studies also show that adding memory-augmented representations helps planners stay coherent over longer sequences. Together, these advances make it possible to build scalable data science workflows that remain aligned with the original goals across many steps \cite{cao2025large}. Together, these advances can make it possible to build scalable data science workflows that remain aligned with the original goals across many steps.

\subsection{Security, Privacy, and Compliance}

Over 90\% of surveyed systems lack explicit mechanisms for security, privacy, and compliance, despite operating on sensitive enterprise data. 
Despite handling sensitive enterprise data, most agents lack privacy-preserving training and secure tool execution. Techniques like differential privacy and federated learning~\cite{yu2021differentially, li2020review} are underused, while threats such as prompt injection~\cite{liu2023agentbench}, membership inference~\cite{hu2022membership}, and cross-modal jailbreaks~\cite{yang2024swe} remain largely unaddressed. Data leakage and the absence of audit trails are especially critical in regulated domains such as finance and healthcare. To address these issues, future systems should employ policy-as-code governance, enforce compliance-aware logging, and adopt secure execution sandboxes to minimize attack surfaces. Techniques such as differential privacy, federated learning, and adversarial testing can strengthen resilience against leakage and manipulation, while integration of auditability and traceability features will ensure legal accountability and regulatory readiness
~\cite{raza2025trism, sculley2015hidden}. In this regard, frameworks like AgentSafe~\cite{mao2025agentsafe} could also be useful. 

%Future systems must integrate DP aware parameter efficient fine tuning, encrypted memory modules, policy based access control, and adversarial training across both language and vision modalities.

%Risks include prompt injection, data leakage, and the absence of audit trails, which are especially critical in regulated domains such as finance and healthcare. To address these, future systems should employ policy-as-code governance, enforce compliance-aware logging, and adopt secure execution sandboxes to minimize attack surfaces. Techniques such as differential privacy, federated learning, and adversarial testing can strengthen resilience against leakage and manipulation, while integration of auditability and traceability features will ensure legal accountability and regulatory readiness.  
%~\cite{raza2025trism, sculley2015hidden}.

\subsection{Trustworthiness, Reliability, and Alignment}
\label{sec:trust-future}

Ensuring safety, trust, and alignment remains a fundamental barrier to deploying DS agents in real-world high-stakes domains. Current systems can plan, invoke external tools, and process sensitive data, yet they rarely incorporate mechanisms to guarantee fairness, transparency, or interpretability. As these agents transition from controlled benchmarks to enterprise settings, their socio technical risks become more pronounced. We identify five pressing challenge areas and outline future research directions.
% \vspace{-0.5\baselineskip}
\paragraph{\textbf{Building Trust and Transparency.}}
Most agents operate as opaque black boxes, offering little visibility into their plans, intermediate reasoning, or tool use decisions. This lack of transparency undermines user confidence and complicates debugging~\cite{plaat2025agentic}. Future systems should adopt transparency by design principles, including auditable logs of decisions, structured chain of thought traces, and interactive plan or code previews. Embedding human in the loop checkpoints throughout the pipeline, where analysts approve critical actions, can further ensure accountability~\cite{wu2023autogen}.
% \vspace{-0.5\baselineskip}
\paragraph{\textbf{Ensuring Reliability and Mitigating Hallucination.}}

Ensuring safety and reliability in DS Agents is critical because these systems face unique challenges, such as propagating biased data, generating incorrect or misleading analyses, exposing sensitive information, and making autonomous decisions that may have unintended consequences. Since LLMs can produce outputs that appear highly convincing even when factually wrong, the risk of misuse or overreliance is significant. To address these issues, robust safety mechanisms should be implemented, including rigorous validation of outputs against trusted data sources, as well as the use of guardrails to prevent access to sensitive information. Additionally, incorporating human-in-the-loop oversight and continuous monitoring can help detect harmful patterns early, ensuring that these agents operate reliably, ethically, and securely within real-world data science workflows. %Moreover, bias in data can propagate through agentic workflows, yet few current systems attempt bias detection or mitigation. Early frameworks like BiasInspector~\cite{li2025biasinspector}, SCALE~\cite{zhaoscale}, and ALI Agent~\cite{wang2024ali} show potential but not extensively used in data science pipelines. Future research should build and integrate automated bias audits into model. %and visualization stages, and develop plug and play fairness modules for multi agent pipelines.

%Bias in data or learned behavior can propagate through agentic workflows, yet few current systems attempt bias detection or mitigation. Early frameworks like BiasInspector~\cite{li2025biasinspector}, SCALE~\cite{zhaoscale}, and ALI Agent~\cite{wang2024ali} show potential but remain siloed from data science pipelines. Future research should build fairness aware datasets, integrate automated bias audits into model and visualization stages, and develop plug and play fairness modules for multi agent pipelines.

%\paragraph{Strengthening Privacy and Security.}
%Most agents lack privacy preserving training and secure tool execution, despite handling sensitive enterprise data. Techniques like differential privacy and federated learning~\cite{yu2021differentially, li2020review} are underused, while threats such as prompt injection~\cite{liu2023agentbench}, membership inference~\cite{hu2022membership}, and cross modal jailbreaks~\cite{yang2024swe} remain largely unaddressed. Future systems must integrate DP aware parameter efficient fine tuning, encrypted memory modules, policy based access control, and adversarial training across both language and vision modalities.

%\paragraph{Improving Robustness and Adversarial Resilience.}

% \vspace{-0.5\baselineskip}
\paragraph{\textbf{Achieving Human Aligned Behavior.}}
Misaligned behavior such as generating misleading analyses or unsafe tool calls remains a core risk. While methods like RLHF, DPO, and GRPO~\cite{ouyang2022training, rafailov2023direct, shao2024deepseekmath} offer alignment pathways, they remain unstable and underexplored for end to end workflows. Future work should focus on scalable alignment using expert preference datasets~\cite{gao2404aligning, chuang2024beyond}, modular oversight mechanisms including plan approval and feedback integration, and continual post-deployment preference tuning.  Staged validation pipelines such as AutoML Agent~\cite{trirat2024automl} suggest promising directions to achieve this. In particular, future DS agents should place greater emphasis on lightweight preference-based alignment methods such as DPO and GRPO, which can help ensure that tasks like visualization quality, model selection, and feature engineering are more consistent with human judgment and domain-specific expectations.

Overall, realizing trustworthy and human aligned DS agents requires treating these concerns not as add ons but as first class design objectives, embedded from dataset curation through deployment governance.

\subsection{Robustness and Generalizability,}

Current DS agents perform well on curated benchmarks but often fail under distribution shifts such as minor schema changes, API updates, unseen visualization tasks, or adversarial inputs, exposing their brittleness in diverse production workflows~\cite{plaat2025agentic, zhang2024benchmarking}. %Future work should combine redundant multi-agent ensembles, constrained decoding, post-generation audits such as SQL or code safety checks, and continuous self-healing strategies for long-horizon workflows.
To build more resilient DS agents, future work must directly address the brittleness exposed by real-world distribution shifts. To counteract failures from minor schema changes and API updates, systems should be designed with {modular skill decomposition}. This allows for targeted updates or fallbacks without retraining the entire agent. To handle unseen visualization tasks and adversarial inputs,  multi-task training \cite{zhang2021survey} and meta learning \cite{hospedales2021meta} for cross-domain transfer, and dynamic adapter routing can help demonstrate generalizability across diverse tasks. By integrating these strategies, agents can sustain robustness not only in controlled benchmarks but also in real-world data science workflows spanning exploratory analysis, modeling, and multimodal reasoning.

%, including exploratory analysis, modeling, and multimodal reasoning

\subsection{Benchmarking and Evaluation}
Existing benchmarks largely focus on isolated tasks, such as SQL generation, visualization, model training, and data cleaning, while neglecting end-to-end workflows that include business understanding, multimodal reasoning, and insight generation \cite{lai2025kramabench,jin2025elt}. This narrow scope limits their ability to capture the complexity of real-world data science pipelines.
To close this gap, future datasets and protocols should evaluate the full lifecycle of data science, with process-oriented metrics that measure error recovery when APIs or schemas shift, scalability under growing workloads, and explainability for human stakeholders. Beyond static correctness checks, benchmarks must incorporate simulation-based stress testing, outcome-grounded verification tied to downstream goals, and trust-aligned assessments that reflect human-centric requirements. Only then can evaluations meaningfully mirror the robustness and adaptability needed in production data science environments~\cite{sahu2024insightbench, jin2025elt, zhang2024benchmarking}.

\subsection{Scalability and Efficiency}
High computational costs and latency remain major barriers to deploying DS agents at scale \cite{belcak2025small}. Many systems require large models for tasks such as multimodal reasoning, orchestration, or long-horizon planning, which limits their practicality in production. Frequent human oversight can further reduce efficiency, especially in high-stakes applications. To address these, future work should prioritize not just lightweight architectures \cite{fu2024tiny} and parameter-efficient fine-tuning \cite{han2024parameter}, but also specific techniques like model quantization \cite{rokh2023comprehensive}, pruning \cite{cheng2024survey}, and knowledge distillation \cite{gou2021knowledge}. These methods should be combined with runtime optimizations like caching, batching, and adaptive computation strategies that can run efficiently on less powerful edge devices. Confidence thresholds and escalation mechanisms can also ensure that agents remain both efficient and reliable while reserving human intervention for only the most critical cases.

\subsection{Societal, Ethical, and Economic Challenges}

The growth of AI agents in data science presents both major opportunities and significant risks. While these agents promise to accelerate workflows and democratize access to analytics, they also risk displacing routine data science tasks, reshaping professional roles, and widening existing skill gaps. In this way, the growing adoption of agentic data science systems raises profound societal, ethical, and economic challenges \cite{santoni2024artificial,shao2025future}. To ensure responsible adoption, agents should be positioned as copilots \cite{hosseini2025role} that augment rather than replace human expertise, with an emphasis on transparency and accountability. Broader engagement with policymakers, educators, and industry stakeholders is needed to establish guidelines that balance innovation with societal readiness.

%The rise of agentic data science systems raises concerns about their impact on employment, professional roles, and ethical responsibility. While these agents can accelerate workflows and democratize access to analytics, they also risk displacing routine data science tasks, creating fears of job loss and widening skill gaps. To ensure responsible adoption, agents should be positioned as copilots that augment rather than replace human expertise, with an emphasis on transparency, accountability, and equitable access to benefits. Broader engagement with policymakers, educators, and industry stakeholders is needed to establish guidelines that balance innovation with societal readiness.

\subsection{Multimodal Understanding and Reasoning}
Modern data science workflows require reasoning across diverse modalities, including charts, tables, code, and text. Yet most current agents struggle with cross-modal grounding and semantic consistency, leading to errors in tasks such as chart interpretation, table summarization, or code–data integration. High token costs for large tables and high-resolution visual inputs further exacerbate these limitations. Promising directions include efficient multimodal architectures and visual reasoning approaches like Visualization-of-Thought \cite{li2025imaginereasoningspacemultimodal}, which integrate intermediate visual steps into the reasoning process. Future benchmarks should evaluate cross-modal grounding, interpretability, and consistency in visual outputs to ensure agents can handle the multimodal nature of real-world data science workflows~\cite{text2vis2025, chartqapro}.

\section{Conclusion}
\label{sec:conclusion}

This survey presents the first comprehensive study of LLM-based agents for data science, offering a lifecycle aligned taxonomy and a systematic evaluation of 45 representative systems. We not only mapped these agents across the stages of the data science workflow but also analyzed what they actually do within each stage, identifying critical trends, strengths, and limitations. We further examined five cross cutting design dimensions including reasoning and planning, modality integration, tool orchestration, learning and alignment, and trust and safety, providing a deeper view into how agentic systems operate in practice. By consolidating a rapidly growing but fragmented body of work, this survey delivers both a structured taxonomy and a critical analysis of gaps that remain unaddressed.

Our analysis reveals several critical trends that we can directly relate to our research questions. \textbf{RQ1:} Current agents concentrate on intermediate stages such as exploratory data analysis and model building, with stage-level synthesis showing strengths in visualization creation and code-based data analysis but persistent gaps in business goal translation, enterprise data acquisition, deployment, and standardized reporting.
\textbf{RQ2:} Multimodal reasoning, tool orchestration, and long-horizon context retention remain key technical bottlenecks; most systems rely on prompt-driven or hierarchical planning, with only emerging and still fragile attempts at multi-agent collaboration, reflection, and adaptive orchestration. \textbf{RQ3:} Trust, safety, and governance mechanisms are largely absent, as agents rarely include formal safeguards for fairness, privacy, explainability, and robustness—leaving them vulnerable to risks such as hallucination, bias propagation, and security vulnerabilities. Existing oversight is limited to isolated human-in-the-loop mechanisms. \textbf{RQ4:} Promising future directions include advancing multimodal grounding, establishing lifecycle-spanning benchmarks that assess end-to-end workflows rather than narrow subtasks, exploring lightweight RL-based alignment methods, and embedding trust, fairness, privacy, and explainability as core design principles. Together, these insights highlight both the progress achieved and the substantial gaps that must be addressed for DS agents to be reliable in real-world, high-stakes settings.

Looking forward, these gaps present critical opportunities for advancing the field. Future research should prioritize modular and generalizable agents that can flexibly adapt across tasks and modalities, supported by standardized benchmarks that evaluate not only final outcomes but also the reasoning processes across the entire lifecycle. Lightweight alignment approaches such as DPO, GRPO, and AgentTuning—together with preference-driven reinforcement learning can foster safer and more human-consistent behavior. Equally important, explainability, privacy, and compliance must be embedded as first-class design principles rather than afterthoughts, ensuring that DS agents can be trusted in high-stakes domains. Ultimately, building reliable, transparent, and human-aligned agents requires weaving these principles into every stage of the lifecycle—from dataset curation to deployment governance—so that the next generation of data science agents not only automates analytics but also democratizes data-driven decision-making in a trustworthy, responsible, and transformative way.

\section*{Acknowledgments}
We thank all contributors and reviewers for their valuable feedback and support.

\bibliographystyle{ACM-Reference-Format}
\renewcommand{\bibfont}{\fontsize{5pt}{6pt}\selectfont}
\bibliography{sample-base}

% \appendix
% \section{Taxonomy of Agentic Data Science Systems}
% % Insert taxonomy table or figure here

% \section{Example Agent Workflow}
% % Insert sample workflow or visual logs of reasoning steps

% \section{Extended Benchmark Comparison}
% % Add any tabular benchmarking comparisons

\end{document}